\pdfoutput=1

\documentclass[11pt]{article}

\usepackage[]{acl}

\usepackage{times}
\usepackage{latexsym}

\usepackage[T1]{fontenc}

\usepackage[utf8]{inputenc}

\usepackage{graphicx}
\usepackage{refstyle}
\usepackage{amsmath}
\usepackage{booktabs}
\usepackage{caption}
\usepackage{subcaption}
\usepackage{soul}
\usepackage{multirow}
\usepackage{longtable}
\usepackage[perpage]{footmisc}
\graphicspath{{figs/}}
\usepackage{xspace}
\newcommand*\methodFont{\textsl}
\newcommand*\CEO{\methodFont{CEO}\xspace}

\usepackage{xcolor}
\definecolor{step1}{HTML}{bf9000}
\definecolor{step2}{HTML}{cc0000}
\definecolor{step3}{HTML}{3d85c6}
\definecolor{step4}{HTML}{38761d}
\definecolor{life}{rgb}{0.0,
        0.10980392156862745,
        0.4980392156862745}
\definecolor{conflict}{rgb}{0.6941176470588235,
        0.25098039215686274,
        0.050980392156862744}
\definecolor{personnel}{rgb}{        0.07058823529411765,
        0.44313725490196076,
        0.10980392156862745}
\definecolor{movement}{rgb}{
        0.5490196078431373,
        0.03137254901960784,
        0.0
}
\definecolor{justice}{rgb}{
        0.34901960784313724,
        0.11764705882352941,
        0.44313725490196076
}
\definecolor{contact}{rgb}{
        0.34901960784313724,
        0.1843137254901961,
        0.050980392156862744
}
\definecolor{business}{rgb}{
        0.6352941176470588,
        0.20784313725490197,
        0.5098039215686274
}
\definecolor{transaction}{rgb}{
        0.23529411764705882,
        0.23529411764705882,
        0.23529411764705882
}

\definecolor{snsblue}{HTML}{1f77b4}
\definecolor{snsorange}{HTML}{ff7f0e}
\definecolor{snsgreen}{HTML}{2ca02c}
\definecolor{snsred}{HTML}{d62728}
\definecolor{snspurple}{HTML}{9467bd}
\definecolor{snspink}{HTML}{e377c2}

\usepackage{cleveref}
\crefformat{section}{\S#2#1#3}
\crefformat{subsection}{\S#2#1#3}
\crefformat{subsubsection}{\S#2#1#3}
\crefrangeformat{section}{\S\S#3#1#4 to~#5#2#6}
\crefmultiformat{section}{\S\S#2#1#3}{ and~#2#1#3}{, #2#1#3}{ and~#2#1#3}
\Crefformat{figure}{#2Fig.~#1#3}
\Crefmultiformat{figure}{Figs.~#2#1#3}{ and~#2#1#3}{, #2#1#3}{ and~#2#1#3}
\Crefformat{table}{#2Tab.~#1#3}
\Crefmultiformat{table}{Tabs.~#2#1#3}{ and~#2#1#3}{, #2#1#3}{ and~#2#1#3}
\Crefformat{appendix}{#2Appx.~\S#1#3}
\crefformat{algorithm}{Alg.~#2#1#3}



\title{\emph{CEO}: \underline{C}orpus-based Open-Domain \underline{E}vent \underline{O}ntology Induction}

\author{Nan Xu$^\diamondsuit$, Hongming Zhang$^\spadesuit$, Jianshu Chen$^\spadesuit$\\
$^\diamondsuit$University of Southern California, $^\spadesuit$Tencent AI Lab, Seattle\\
\texttt{$^\diamondsuit$nanx@usc.edu, $^\spadesuit$\{hongmzhang,jianshuchen\}@global.tencent.com
}}

\begin{document}
\maketitle
\begin{abstract}

Existing event-centric NLP models often only apply to the pre-defined ontology, which significantly restricts their generalization capabilities.
This paper presents \CEO, a novel Corpus-based Event Ontology induction model to relax the restriction imposed by pre-defined event ontologies. 
Without direct supervision, \CEO leverages distant supervision from available summary datasets to detect corpus-wise salient events and exploits external event knowledge to force events within a short distance to have close embeddings. 
Experiments on three popular event datasets show that the schema induced by \CEO has better coverage and higher accuracy than previous methods.
Moreover, \CEO is the first event ontology induction model that can induce a hierarchical event ontology with meaningful names on eleven open-domain corpora, making the induced schema more trustworthy and easier to be further curated. We release our dataset, codes, and induced ontology.~\footnote{\url{https://sites.google.com/view/ceoeventontology}}
\end{abstract}

\section{Introduction}
\begin{figure}[t!]
\centering
\includegraphics[width=\linewidth]{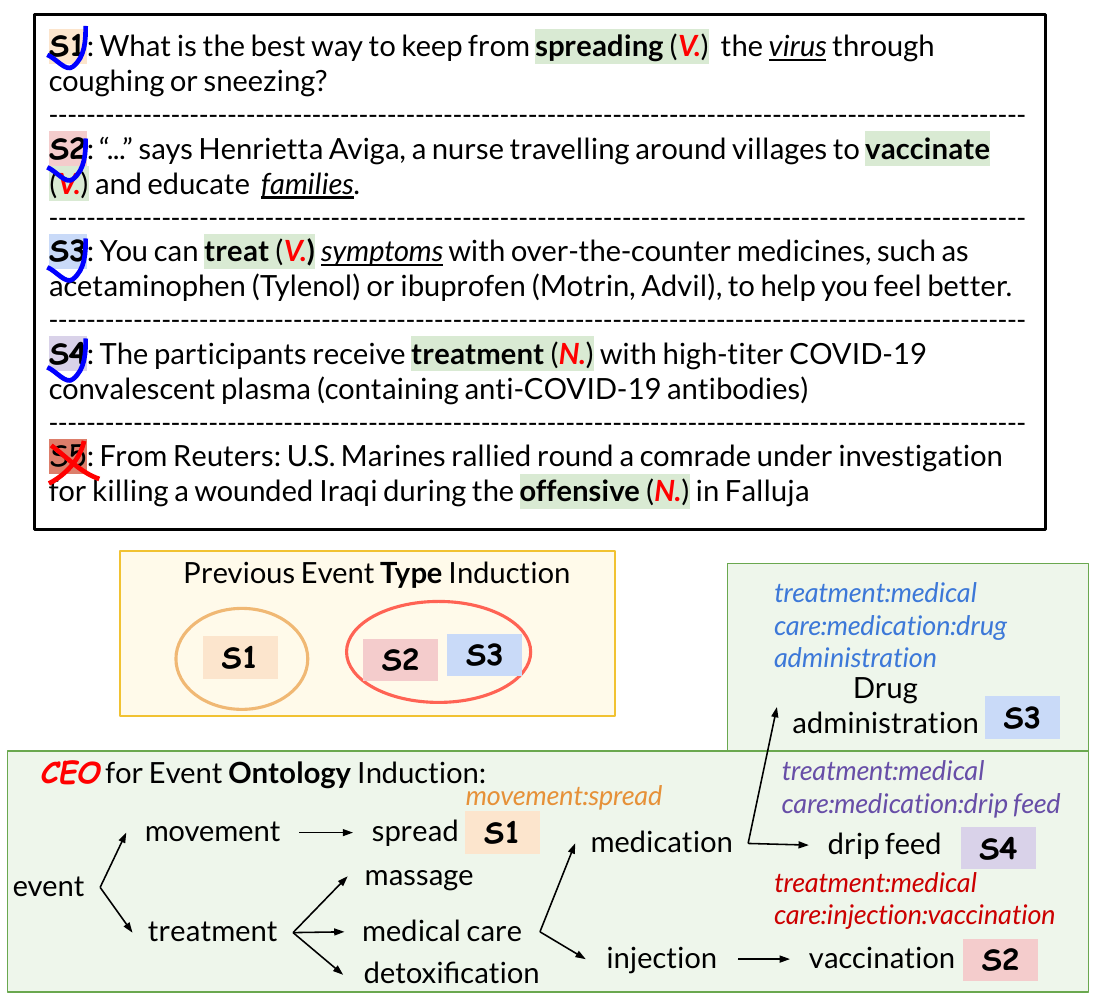}
\caption{Instances from Covid-19 corpus with event type induced by previous work and ontology induced by \CEO. The non-salient event \emph{treatment}in \emph{S4} is disregarded while others are preserved. Event \textbf{type} induction only identifies events triggered by verbs (\emph{S1}, \emph{S2}, \emph{S3}) but not nouns (\emph{S4}), and arranges events into simple clusters. \CEO recognizes both verb- and noun-triggered events, induces tree-structure ontology and provides concrete names.}
\label{fig:overview}
\vspace{-.1in}
\end{figure}
Extracting and understanding real-world events described in the text are crucial information extraction tasks that lay the foundations for downstream NLP applications~\cite{chen-etal-2021-event,zhang2020diagnostic,fung-etal-2021-infosurgeon}.
However, existing event-related studies are mostly restricted by the pre-defined ontology~\cite{zhang-etal-2022-zero,guzman-nateras-etal-2022-cross}. Even for the zero-shot setting, models still need a pre-defined ontology for inference~\cite{huang-ji-2020-semi,edwards2022semi}.

To address this limitation, the previous work~\cite{shen-etal-2021-corpus} proposed the \emph{event type induction} task, which automatically induces event ontology from documents.
However, previous work only covers verbal events while ignoring the nominal ones.
Moreover, it can only induce the flat ontology, which is not enough to cover the rich hierarchical ontology structure defined by humans.
Last but not least, the induced ontology only contains type ids, making it hard to be verified and curated by users.

This paper introduces a new Corpus-based open-domain Event Ontology induction strategy (\CEO).
As demonstrated in Figure~\ref{fig:overview},
\CEO covers both verbal and nominal events and leverages external summarization datasets to detect salient events better.
On top of that, \CEO is also capable of inducing hierarchical event ontology with the help of a word sense ontology tree defined in WordNet~\cite{fellbaum2010wordnet}.
To enhance the faithfulness of induced ontology and facilitate future curation, \CEO generates a meaningful name for each induced event type in the induced ontology.


In the proposed \CEO strategy, we make two key technical contributions to better learn from open-domain events. 
The first technical contribution is 
corpus-wise salient event detection
with distant supervision from available summary datasets. 
Following the assumption that summaries written by humans are likely to include events about the main content~\cite{liu-etal-2018-automatic,jindal-etal-2020-killed}, we consider events mentioned both in summary and body text as salient while those only mentioned in the body text as non-salient. 
To obtain corpus-wise key events, we fine-tune a Longformer-based model~\cite{beltagy2020longformer} to classify whether the identified events are salient or not given rich context.

The second contribution is exploiting external event knowledge for hierarchical open-domain event ontology inference. Specifically, we leverage the word sense
ontology (i.e., the hypernym/hyponym relationships) trees in WordNet~\cite{fellbaum2010wordnet} to improve event representations.
We propose to train an autoencoder model~\cite{domingos2015master} to compress the original event representations in the latent space, where information is preserved by minimizing the reconstruction error. We further utilize a triplet loss~\cite{balntas2016learning} to regularize the compressed embeddings, so that event pairs with senses in a short distance in the WordNet ontology tree are much closer (i.e., anchor and positive events) compared with those far away from each other (i.e., anchor and negative events). After training event data from both WordNet and the studied corpus with ontology supervision from the former, events with close compressed embeddings in the latter are expected to have short distances in the ontology tree.

In summary, we propose an effective strategy, \emph{CEO}, to extract and understand corpus-based open-domain events. 
Experiments on three popular event datasets show that the proposed \emph{CEO} could consistently induce accurate and broad-coverage event ontology without direct supervision.
Moreover, to the best of our knowledge, \CEO is the best model that could induce a hierarchical event ontology with meaningful names. We also perform event ontology induction on $11$ open-domain news corpus such as \emph{abortion, LGBT} and demonstrate the broad application of \CEO.

\section{Related Work}
\textbf{Event Extraction}\quad Given a set of pre-defined types and annotated samples, event extraction is typically cast as a multi-class classification task, where event types and argument roles are predicted into one of target types~\cite{lin-etal-2020-joint}. Recently, semantic meanings of event and argument types have gained much attention to capture correlations between event mentions and  types~\cite{wang-etal-2022-query,hsu-etal-2022-degree}. 
~\\\textbf{Semi- and Un-supervised Event Type Induction}
\quad To classify constantly emerging events of new types without annotations in an existing domain, semi-supervised learning approaches such as Vector Quantized Variational Autoencoder~\cite{huang-ji-2020-semi} and contrastive learning~\cite{edwards2022semi,zhang-etal-2022-zero} have been introduced. ETypeClus~\cite{shen-etal-2021-corpus} proposed to perform event type induction under the unsupervised setting, where neither annotations nor event types are used. 
Different from unutterable event clusters induced by ETypeClus, \emph{CEO} infers underlying event type ontology including interpretable type for each mention in diverse granularities. 
\begin{figure*}[t!]
\centering
\includegraphics[width=.8\textwidth]{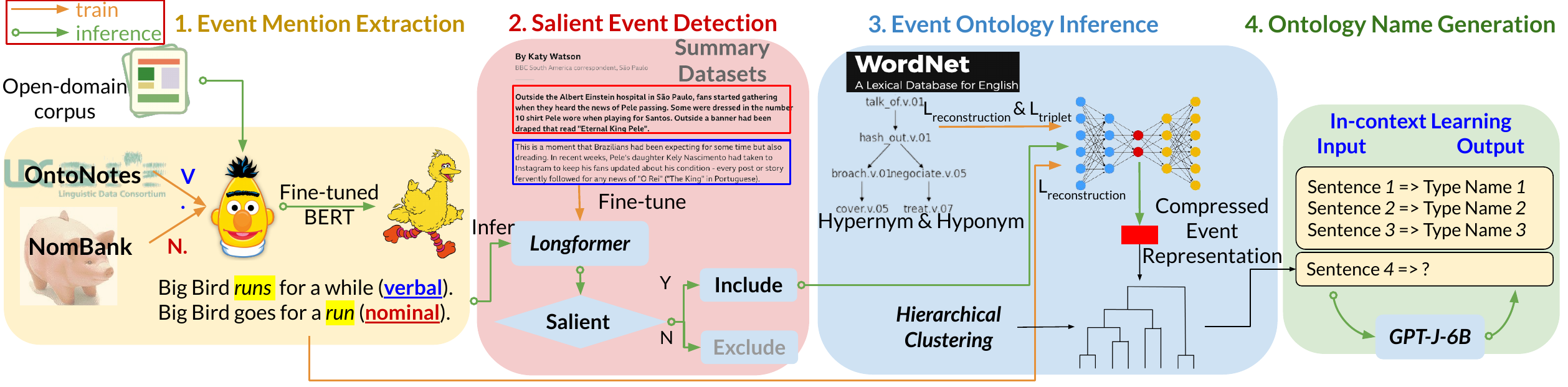}
\caption{Framework of the proposed \CEO. \textcolor{step1}{\emph{Step 1}}: extract events triggered by nouns or verbs; \textcolor{step2}{\emph{Step 2}}: preserve salient events with distant supervision from summaries; \textcolor{step3}{\emph{Step 3}}: improve event representations for hierarchical clustering with external event knowledge from WordNet; \textcolor{step4}{\emph{Step 4}}: generate event type names with in-context learning.}
\label{fig:framework}
\vspace{-.2in}
\end{figure*}
\section{Problem Definition}
Since the majority of events are triggered by \textbf{verbal} and \textbf{nominal} predicates along with relevant arguments, we denote an event mention by $<$\emph{subject}, \emph{predicate}, \emph{object}$>$. For each corpus, event mentions highly relevant to its topic are considered as \textbf{salient} and constitute the extraction targets.  
To understand semantic relations between events, we aim at inducing a hierarchical event type \textbf{ontology} with a tree structure, where leaf nodes represent single event mentions while internal nodes are subclusters of events.
\paragraph{Task Definition.} Given a corpus of $N$ sentences $\mathcal{C}=\{S_1,\dots,S_N\}$,
\emph{event ontology induction} 1) firstly extracts salient event mentions, e.g., $m_{ij}$ for $j$-th event in $S_i$, 2) then identifies event ontology that well demonstrates correlations among all covered event types, 3) lastly infers event type names withing human readable formats from coarse-to-fine granularity.
\section{CEO}
In \Cref{fig:framework}, we show the overview of the proposed \CEO that extracts (\textcolor{step1}{\emph{Step $1$}} in \Cref{sec:event_mention_extraction}) and represents salient  events (\textcolor{step2}{\emph{Step $2$}} in \Cref{sec:salient_event_detection}) with informative embeddings for ontology structure induction (\textcolor{step3}{\emph{Step $3$}} in \Cref{sec:method_ontology_inference}) and name generation (\textcolor{step4}{\emph{Step $4$}} in \Cref{sec:ontology_name_generation}). 
\subsection{Event Mention Extraction}\label{sec:event_mention_extraction}

We take advantage of event trigger-annotated datasets, OntoNotes~\cite{pradhan2013towards} and NomBank~\cite{meyers-etal-2004-nombank}, for verb- and noun-triggered event information extraction, respectively.  Concretely, we adopt a two-stage process for event information extraction: \emph{1) event trigger detection:} we follow the practice in~\cite{shen-etal-2021-corpus} to extract verbal tokens identified by the dependency parser as the verbal event trigger; since nouns play much more diverse roles in sentences besides predicates, we cast the nominal predicate detection as a binary classification task and fine-tune the BERT~\cite{devlin-etal-2019-bert} model to identify nouns labeled as event triggers in NomBank\footnote{NomBank is an open-domain dataset with broad coverage that considers nouns in Wall Street Journal Corpus of the Penn Treebank~\cite{garofolo1993csr}.}. \emph{2) joint training for event-relevant information learning:} with the identified event triggers, we follow the work for semantic role labeling~\cite{shi2019simple,SRLEnglish2021}, where the vanilla BERT  model is connected with two linear layers, one for argument classification and the other for predicate sense disambiguation. The extracted event information from \CEO, including event trigger tokens, their semantic senses, and accompanying argument tokens, comprehensively describes different perspectives of events. 

\subsection{Salient Event Detection}\label{sec:salient_event_detection}

\begin{table*}[t!]
\centering
\resizebox{\textwidth}{!}{%
\begin{tabular}{@{}l@{}}
\toprule
\textbf{Title}: Metro Briefing | New York : Brooklyn : Charter Review Meeting Disrupted .\\ \midrule
\begin{tabular}[c]{@{}l@{}}\textbf{Summary}: First public hearing of \emph{\textcolor{red}{Charter}} Revision  \emph{\textcolor{red}{Commission}} is disrupted by protesters Daniel Cantor and Arron Schildkrout, \\ who oppose New York City Mayor Michael R Bloomberg's plan to institute nonpartisan  \emph{\textcolor{red}{elections}} ( S )\end{tabular} \\\midrule
\begin{tabular}[c]{@{}l@{}}\textbf{Body Text}: The first public hearing of Mayor Michael R. Bloomberg's  \emph{\textcolor{red}{Charter}} Revision  \emph{\textcolor{red}{Commission}} was disrupted last night by \\ protesters, and two men were  \emph{\textcolor{blue}{arrested}}. Opponents of the mayor's plan to \emph{\textcolor{blue}{establish}} nonpartisan  \emph{\textcolor{red}{elections}} burst into the Fire \\ Department's headquarters in Brooklyn, where the hearing was held, and \emph{\textcolor{blue}{chanted}}, '' \emph{\textcolor{blue}{Change}} the mayor, not the \emph{\textcolor{red}{charter}}. '' \\ Two men, Daniel Cantor, 47, of Brooklyn, and Arron Schildkrout, 22, of Watertown, Mass., were \emph{\textcolor{blue}{arrested}} and \emph{\textcolor{blue}{charged}} with ...
\end{tabular} \\ \bottomrule
\end{tabular}%
}
\caption{Instance sampled from NYT Corpus. Event triggers in the body text are marked in \emph{italic}. Events concurrently mentioned in summary and body text are deemed salient and in \emph{\textcolor{red}{red}}, while others are non-salient in \emph{\textcolor{blue}{blue}}.}
\label{tab:summary_instance}
\vspace{-.1in}
\end{table*}

Aimed at only extracting events salient to the given corpus, prior work~\cite{shen-etal-2021-corpus} adopted the TF-IDF idea and defined the event salience by comparing the frequency of trigger words in the studied corpus against a  general-domain corpus. We argue that such a rough criterion disregards contextual information of event triggers and is prone to cause massive \emph{false negatives}.\footnote{For instance, the surface pattern of a trigger word could be rarely observed, but its semantic relevance to the corpus theme might be very high.} Instead, we detect salient events based on the semantic and contextual information of predicates. As shown in \Cref{tab:summary_instance}, we propose to leverage distant supervision from summarization datasets,~\footnote{Different from prior work that focuses on either solving summarization task with external knowledge~\cite{zhang-etal-2023-enhancing} or reformulating another task as summarization~\cite{lu-etal-2022-summarization}, we leverage summarization datasets and models to extract salient events from documents.} 
following the assumption that an event is considered salient if a summary written by a human tends to include it~\cite{liu-etal-2018-automatic,jindal-etal-2020-killed}. To consider a wide window of context, we fine-tune the Longformer~\cite{beltagy2020longformer} model to perform binary classification: given contexts and trigger words, predict the events as salient if they appear in summary as well. For open-domain event salience inference, we provide the event sentence with context and obtain its corresponding salience score.
\subsection{Event Ontology  Inference}\label{sec:method_ontology_inference}
With all kinds of event-centric information for salient events, we can infer the corpus-level event ontology by incorporating the learned informative event embeddings into a wide range of off-the-shelf hierarchical clustering models (discussed in  \Cref{sec:hierarchical_clustering}). For individual event mentions, we average over the following embeddings as the final comprehensive event representations: \emph{1) contextualized embeddings} for tokens at positions predicted as the predicate, subject, and object; \emph{2) event sentence embeddings} represented by Sentence-BERT~\cite{reimers-gurevych-2019-sentence}; \emph{3) predicate sense embeddings} composed of definition sentence representations from Sentence-BERT and contextualized token embeddings for predicate positions from example sentences. 

Although there is no extra knowledge about the actual event ontology of the studied open-domain corpus, we find that the explicit hypernym/hyponym relationships among the verb synsets in WordNet~\cite{fellbaum2010wordnet} can provide concrete guidance for the hierarchical event ontology\footnote{The latest WordNet contains 13,650 verb synsets.}. To further improve event embeddings, we exploit the event ontology in WordNet by augmenting the standard autoencoder with an additional contrastive loss. We first assume that events within a short distance from each other in the ontology tree should be semantically similar and close in the latent space of the autoencoder (see \Cref{sec:autoencoder_design} for distance computation and \Cref{fig:external_wordnet} for visualization). We then utilize the following loss function to augment the reconstruction loss for optimizing the autoencoder parameters\footnote{As demonstrated in \Cref{fig:framework} and \Cref{fig:external_wordnet}, to avoid distribution shift, events predicted from the studied corpus is also used for  reconstruction loss besides those annotated in WordNet, but only the latter is available hence used for triplet loss.}: $L_{\text{triplet}}(i,p,n)=\max\{d(\mathbf{e_i},\mathbf{e_p})-d(\mathbf{e_i},\mathbf{e_n})+\text{margin},0\}$,
where $i$, $p$ and $n$ are anchor, positive, and negative events, $\mathbf{e_i}$, $\mathbf{e_p}$ and $\mathbf{e_n}$ are their representations in the latent space, $d$ denotes the Euclidean distance. Compressed vectors in the latent space are adopted for ontology inference.
\subsection{Ontology Name Generation}\label{sec:ontology_name_generation}
From the bottom leaf layer to the top root node in the learned ontology tree, diverse event instances are clustered according to different levels of similarities. Motivated by the in-context learning capacity of pre-trained language models, we randomly sample event instances from other available event datasets as demonstrations 
(see an in-context learning example in \Cref{tab:name_generation_prompt}).
For internal node name generation, the token probability distribution of event type names is averaged over all included events and the most likely is selected.

\section{Experiments}
In this section, we firstly introduce the utilized event datasets (\Cref{sec:dataset}) and then quantitatively evaluate the ontology (\Cref{sec:hierarchical_clustering}) and name (\Cref{sec:name_generation}) induction quality of \CEO. 
Then we evaluate the effectiveness of different techniques incorporated in \CEO (\Cref{sec:ablation}) via the ablation study.
Lastly, we apply \CEO to perform ontology induction on eleven open-domain corpora (\Cref{sec:allsides_exp}) to demonstrate its effectiveness in real applications.
\subsection{Datasets}~\label{sec:dataset}
\begin{table}[t!]
\resizebox{\columnwidth}{!}{%
\begin{tabular}{@{}lcccc@{}}
\toprule
Dataset  & \#Docs & \begin{tabular}[c]{@{}c@{}}\#Event\\ Mentions\end{tabular} & \begin{tabular}[c]{@{}c@{}}\#Event\\ Types (Ontology)\end{tabular}&\begin{tabular}[c]{@{}c@{}}\%Predicates\\ Noun/Verb\end{tabular} \\ \midrule
ACE 2005 & 599    & 5,349                                                      & 33 (2 levels)  &43.73/46.34                                                    \\
MAVEN    & 4,480  & 118,732                                                    & 168 (4 levels) &28.60/64.23                                                    \\
RAMS     & 3,993  & 9,124                                                      & 139 (3 levels)&39.99/55.45                                                         \\ \bottomrule
\end{tabular}%
}
\caption{Statistics of studied event datasets show nouns are as important as verbs in expressing events.
}
\label{tab:statistics}
\vspace{-.2in}
\end{table}
We summarize statistics of utilized event datasets in \Cref{tab:statistics} and visualize their corresponding ontologies in \Cref{fig:event_ontologies}. \textbf{ACE2005}~\cite{doddington-etal-2004-automatic} is the widely used English event dataset 
with its event schema organized by a 2-level hierarchy: five types of general events, each with $1{\sim}13$ subtypes included. 
\textbf{MAVEN}~\cite{wang-etal-2020-maven} is a massive general domain event detection dataset 
with its event types manually derived from 
the linguistic resource FrameNet~\cite{baker1998berkeley} following a 4-layer tree-structure.
\textbf{RAMS}~\cite{ebner-etal-2020-multi} employs a three-level hierarchical event ontology
with all types annotated according to a manually constructed mapping.

\begin{table}[t!]
\resizebox{\columnwidth}{!}{%
\begin{tabular}{@{}lcccccc@{}}
\toprule
\multirow{2}{*}{Methods}                                & \multicolumn{2}{c}{ACE2005} & \multicolumn{2}{c}{MAVEN} & \multicolumn{2}{c}{RAMS} \\ \cmidrule(l){2-7} 
 &
  Purity $\uparrow$ &
  \begin{tabular}[c]{@{}c@{}}Cost $\downarrow$\\ ($\times10^9$)\end{tabular} &
  Purity $\uparrow$ &
  \begin{tabular}[c]{@{}c@{}}Cost $\downarrow$\\ ($\times10^{12}$)\end{tabular} &
  Purity $\uparrow$ &
  \begin{tabular}[c]{@{}c@{}}Cost $\downarrow$\\ ($\times10^9$)\end{tabular} \\ \midrule
hkmeans                                                 & .519         & \textbf{1.00}         & .356        & \textbf{4.75}        & .143        & 6.79       \\\midrule
birch                                                   & .242         & 1.49         & .129        & 6.88        & .057        & 8.00       \\\midrule
perch                                                   & .370         & 1.01        & .361        & 4.78        & .154        & 6.84       \\\midrule
ghhc                                                    & .189         & 1.54         & .027        & 7.22        & .019        & 10.3       \\\midrule
HypHC                                                   & .302         & \textbf{1.00}         & .027        & 4.81        & .040        & \textbf{6.75}       \\\midrule
\begin{tabular}[c]{@{}l@{}}ward \\linkage\end{tabular} & \textbf{.556}         & \textbf{1.00}         & \textbf{.457}        & \textbf{4.75}        & \textbf{.220}        & 6.78       \\\bottomrule
\end{tabular}
}
\caption{Performance of our ward linkage and other hierarchical clustering methods evaluated by dendrogram purity and Dasgupta cost. Inferred hierarchical clusters with higher purity ($\uparrow$) and lower cost ($\downarrow$) are more aligned with the ground-truth event ontologies.}
\label{tab:hier_perf}
\vspace{-.2in}
\end{table}
\subsection{Implementation Details}\label{sec:implementation_details}
For event mention extraction (~\Cref{sec:event_mention_extraction}), BERT is fine-tuned for event extraction model on OntoNotes for verbal predicates and Nombank for nominal predicates. For salient event detection (~\Cref{sec:salient_event_detection}), we label events as salient if they also appear in summary; for New York Times, both events in summary and body text are annotated. For event ontology inference (~\Cref{sec:method_ontology_inference}), the encoder layers are [896, 768, 640, 512], while the decoder layers are the reverse for the Autoencoder; the learning rate is $0.005$ and training epochs are $100$. 
\subsection{Evaluations of Event Ontology Induction}
In this section, we evaluate induced event ontologies from two perspectives: mention clustering accuracy and cluster name preciseness.
\subsubsection{Hierarchical Clustering}~\label{sec:hierarchical_clustering}
\textbf{Metrics}\quad We evaluate the quality of inferred hierarchical clusters using the widely-adopted \emph{dendrogram purity}~\cite{heller2005bayesian}, and the more recent \emph{Dasgupta cost}~\cite{dasgupta2016cost}.
Higher purity and lower cost indicate more accurate clustering.
We leave their concrete formulae in \Cref{sec:hierarchical_clustering_metric}.
\begin{figure*}[t!]
\centering
\includegraphics[width=.65\textwidth]{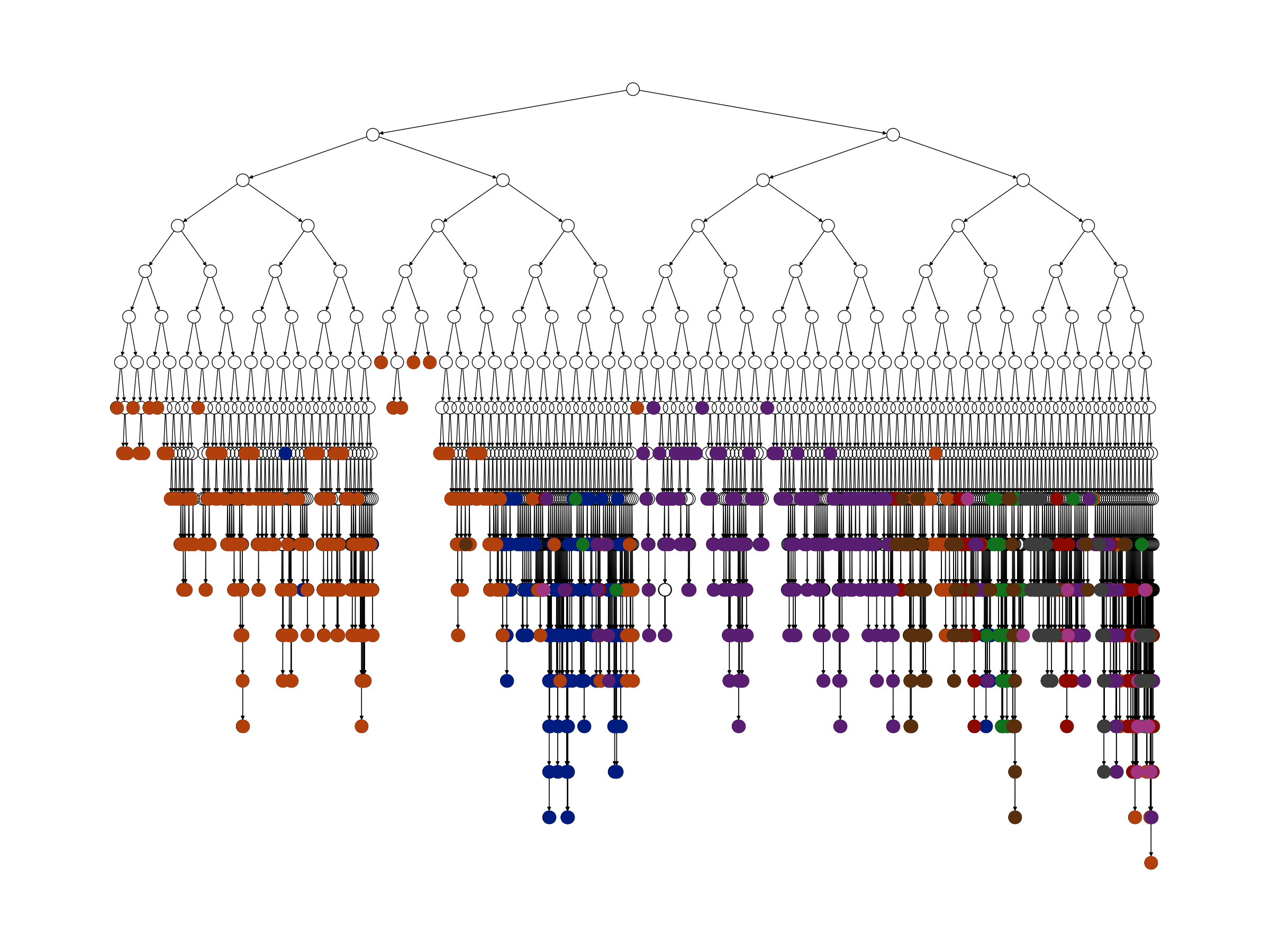}
\vspace{-.5in}
\caption{Event ontology induced by ward linkage on ACE2005. Each leaf node represents one event mention and is colored by its actual coarsest event type: \emph{\textcolor{life}{Life}}, \emph{\textcolor{personnel}{Personnel}}, \emph{\textcolor{justice}{Justice}}, \emph{\textcolor{conflict}{Conflict}}, \emph{\textcolor{transaction}{Transaction}}, \emph{\textcolor{movement}{Movement}}, \emph{\textcolor{contact}{Contact}}, \emph{\textcolor{business}{Business}}. The ontology hierarchies of the other two datasets are visualized in \Cref{fig:datasets_full}.}
\label{fig:ace2005}
\end{figure*}

\paragraph{Baselines}
We perform comprehensive evaluations on discrete optimization methods from two classes: top-down divisive --\emph{Hierarchical Kmeans} and \emph{Birch}~\cite{zhang1997birch}, and bottom-up agglomerative --\emph{Ward Linkage}~\cite{ward1963hierarchical} and \emph{Perch}~\cite{kobren2017hierarchical}. 
Furthermore, we consider recent gradient-based continuous optimization methods which benefit from stochastic optimization:
\emph{gHHC}~\cite{monath2019gradient} and \emph{HypHC}~\cite{chami2020trees}.
\paragraph{Results}
As shown in \Cref{tab:hier_perf}, we adopt \textbf{\emph{ward linkage}} algorithm, which achieves the best performance for ontology induction evaluated by both purity and cost consistently. On MAVEN and RAMS with more complicated event ontologies, the enlarged performance gap is observed between continuous optimization methods and discrete ones. We speculate that hundreds of clusters and input dimensions make it challenging for the continuous approach to outperform discrete methods based on heuristics, which is in contrast to observations reported on small-scale datasets~\cite{monath2019gradient,chami2020trees}.

We further demonstrate the alignment of inferred event ontology with coarsest event type annotations for ACE 2005 in \Cref{fig:ace2005} and the other two datasets in \Cref{fig:datasets_full}. We observe that events of identical coarse-grained types are clustered together compared with those annotated by different labels. In \Cref{fig:ace2005}, the most popular \emph{\textcolor{conflict}{conflict}} events cluster in the left branches while the less popular \emph{\textcolor{justice}{justice}} events gather in the middle branches. 
  
\begin{table*}[t!]
\resizebox{\textwidth}{!}{%
\begin{tabular}{@{}lccccccccc@{}}\toprule
\multirow{2}{*}{Method} & \multicolumn{3}{c}{ACE2005}                   & \multicolumn{3}{c}{MAVEN}                     & \multicolumn{3}{c}{RAMS}                      \\\cmidrule(lr){2-10}
                        & Sim dist $\uparrow$      & rougeL $\uparrow$        & BERTScore $\uparrow$     & Sim dist $\uparrow$      & rougeL $\uparrow$        & BERTScore $\uparrow$     & Sim dist $\uparrow$      & rougeL $\uparrow$        & BERTScore $\uparrow$     \\\midrule
most frequent & .508 & .167 & .869 & .466 & .043 & .836 & .448 & .041 & .849 \\
tf-idf        & .505 & .184 & .869 & .464 & .041 & .835 & .447 & .038 & .849 \\\midrule
topicrank     & .437 & .024 & .824 &   .380   & 0.0     &  .721    & .413 & .006 & .817 \\
textrank      & .418 & .035 & .813 &   .376   & 0.0     &  .724    & .399 & .016 & .811 \\
keybert       & .462 & .072 & .838 &  .427    &   0.0   & .795     & .425 & .014 & .830 \\\midrule
WordNet       & .438 & .055 & .827 & .418 & .006 & .814 & .411 & .003 & .825 \\\midrule
RoBERTa-large & .510 & .191 & .871 & .462 & .041 & .838 & .440 & .027 & .842 \\
GPT-J-6B                & \textbf{.513} & \textbf{.210} & \textbf{.880} & \textbf{.466} & \textbf{.051} & \textbf{.840} & \textbf{.466} & \textbf{.086} & \textbf{.851}\\\bottomrule
\end{tabular}%
}
\caption{Evaluation of type names from our GPT-J-6B and other generation methods for event ontologies. For all metrics, higher scores indicate higher similarity of generated names to the annotated hierarchical event labels.}
\label{tab:name_perf}
\vspace{-.1in}
\end{table*}
\subsubsection{Name Generation}~\label{sec:name_generation}
\textbf{Metrics}\quad
We treat the ground-truth coarse-to-fine label names, $E_r=\{e^i_r|1\leq i\leq n_r\}$ of $n_r$ levels, as an ordered reference. We compare $E_r$ with the generated type names, which are composed of node names from root to leaf in the ontology tree, $E_p=\{e^j_p|1\leq j\leq n_p\}$ of $n_p$ levels. 
We utilize the following metrics:
\emph{1) Sim dist} is self-defined to consider both semantic similarity and granularity difference between each pair of reference $e^i_r$ and generated name $e^j_p$ (see \Cref{sec:name_generation_metrics} for the formula);
\emph{2) Rouge-L}: type names from coarse to fine granularities are combined into a single sentence and Rouge-L score~\cite{lin-2004-rouge} is used to compare the generated against the reference sentence. 
\emph{3) BERTScore}~\cite{zhang2019bertscore}: similar to Rouge-L, the similarity F1 score is computed for token pairs in the generated and reference sentence. 
\paragraph{Baselines}
With clustered events predicted by \CEO, we utilize either statistical strategies -- \emph{Most frequent} and \emph{tf-idf}, or off-the-shelf language models -- \emph{RoBERTa-large}~\cite{liu2019roberta} and \emph{GPT-J-6B}~\cite{wang2021gpt}, to generate cluster names. Keywords extracted by \emph{textrank}~\cite{mihalcea-tarau-2004-textrank}, \emph{topicrank}~\cite{bougouin-etal-2013-topicrank} or \emph{KeyBERT}~\cite{grootendorst2020keybert} are also utilized as cluster names. Besides, we introduce the \emph{wordnet synset} strategy that adopts the least common ancestor hypernym of event triggers~\cite{fellbaum2010wordnet}. We describe more methodology details in \Cref{sec:name_generation_strategy}.
\paragraph{Results}
We evaluate the qualities of our in-context learning \textbf{\emph{GPT-J-6B}} and other name generation strategies and show results in \Cref{tab:name_perf}. 
The language model \emph{GPT-J-6B} achieves the best performance evaluated by three metrics on all studied datasets. Compared with other statistical methods, keyword extraction strategies can hardly extract salient event triggers from thousands of tokens.
Overall, deep language models perform much better than statistical ones.
\paragraph{Human Evaluations}
For each event dataset, we randomly sample $100$ instances and ask annotators to compare type names from \emph{GPT-J-6B} and the 2nd best strategy in~\Cref{tab:name_perf}. As demonstrated in~\Cref{tab:human_eval}, event names generated by \emph{GPT-J-6B} are consistently preferred across three datasets.
\begin{table}[t!]
\centering
\resizebox{.8\columnwidth}{!}{%
\begin{tabular}{@{}lccc@{}}
\toprule
Preference      & ACE2005 & MAVEN & RAMS \\ \midrule
GPT-J-6B better & \textbf{.75}     & \textbf{.58}   & \textbf{.59}  \\
2nd best better & .21     & .30   & .22  \\
Same            & .04     & .12   & .19  \\ \bottomrule
\end{tabular}%
}
\caption{Human preferences on event names generated by GPT-J-6B and 2nd best strategy for each dataset.}
\label{tab:human_eval}
\vspace{-.25in}
\end{table}
\paragraph{Case Study}
We randomly sample three event instances and demonstrate their type names generated from different strategies in \Cref{tab:event_case_study}. For easy instances such as \hyperlink{t1}{\emph{T1}} and \hyperlink{t2}{\emph{T2}}, we observe that statistical strategies are able to produce type names as accurately as pre-trained LMs. However, for the challenging instance \hyperlink{t3}{\emph{T3}}, most generation strategies mistakenly provide descriptions semantically opposite to \emph{robs}, e.g., \emph{lend} and \emph{borrow} from \emph{WordNet Sysnet}.
Only \emph{GPT-j-6B} successfully captures the critical meaning of the event: \emph{attack} and \emph{steal}.
\begin{table*}[t!]
\resizebox{\textwidth}{!}{%
\begin{tabular}{@{}ll@{}}\toprule
Dataset                   & Event Instances and Names                                                                                                                                                                                                                                                                                                                                      \\\midrule
\multirow{2}{*}{ACE2005} & \hypertarget{t1}{\emph{T1:}} Peterson Trial Scott Peterson has been found guilty of \colorbox{yellow}{\emph{murdering}} his wife Laci and their unborn son, and he now faces the death penalty.                                                                                                                                                                                                                       \\
                          & \begin{tabular}[c]{@{}l@{}}\textbf{Gold types}: \textcolor{snsblue}{life:die} \quad Most Frequent: \textcolor{snsorange}{kill:die:murder}\quad TF-IDF: \textcolor{snsgreen}{kill:die:murder}\\ WordNet Synset: \textcolor{snspink}{killing:die:murder}\quad RoBERTa-large: \textcolor{snspurple}{kill:die:murder}\quad GPT-j-6B: \textcolor{snsred}{death:murder}\end{tabular}                                                                                                                                                           \\\midrule
\multirow{2}{*}{MAVEN}   
& \hypertarget{t2}{\emph{T2:}} The robbers attempted to \colorbox{yellow}{\emph{flee}} the scene, Phillips on foot and Matasareanu in their getaway vehicle while continuing to exchange fire with the officers.                                                                                                                                                                                                       \\
                          & \begin{tabular}[c]{@{}l@{}}\textbf{Gold types}: \textcolor{snsblue}{Action:Motion:Self\_motion:Escaping}\quad Most Frequent: \textcolor{snsorange}{attack:meet:send:move:fly:transport:carry}\\ TF-IDF: \textcolor{snsgreen}{become:destroy:receive:occupy:evacuate:flee}\quad WordNet Synset: \textcolor{snspink}{range:destroy:pit:inflict:seize:flee}\\ RoBERTa-large: \textcolor{snspurple}{hold:destroy:receive:occupy:evacuate:flee}\quad GPT-j-6B: \textcolor{snsred}{attack:transport:escape}\end{tabular} \\\midrule
\multirow{2}{*}{RAMS}     & \begin{tabular}[c]{@{}l@{}}\hypertarget{t3}{\emph{T3:}} Corruption in oil production - one of the world's richest industries and one that touches us all through our reliance on petrol - fuels inequality, \colorbox{yellow}{\emph{robs}} \\ people of their basic needs and causes social unrest in some of the world's poorest countries\end{tabular}                                                           \\
                          & \begin{tabular}[c]{@{}l@{}}\textbf{Gold types}: \textcolor{snsblue}{conflict:attack}\quad Most Frequent: \textcolor{snsorange}{urge:donate:lend:borrow:rob}\quad TF-IDF: \textcolor{snsgreen}{urge:donate:lend:borrow:rob}\\ WordNet Synset: \textcolor{snspink}{rede:donate:borrow:rob}\quad RoBERTa-large: \textcolor{snspurple}{urge:donate:end:rob}\quad GPT-j-6B: \textcolor{snsred}{attack:transfer:steal}\end{tabular}\\\bottomrule                                                                                                                       
\end{tabular}}
\caption{Generated names for instances sampled from three event datasets. We mark the predicted \colorbox{yellow}{\emph{predicates}}, while type names are separated by ``:'' and arranged from coarse to fine.}
\label{tab:event_case_study}
\vspace{-.1in}
\end{table*}
\subsection{Ablation Studies}~\label{sec:ablation}
In this section, we showcase the effectiveness of different techniques introduced in \CEO. 
\paragraph{Benefits of Event Embedding}\label{sec:embedding_benefits}
\begin{table}[t!]
\centering
\resizebox{.85\columnwidth}{!}{%
\begin{tabular}{@{}llccc@{}}
\toprule
Predicate                 &     & ACE2005       & MAVEN         & RAMS          \\ \midrule
\multirow{2}{*}{Nominal}  &ETypeClus & -             & -             & -             \\
                          & \CEO       & \textbf{.630} & \textbf{.612} & \textbf{.600} \\\midrule
\multirow{2}{*}{Verbal} & ETypeClus & .713          & .770          & .764          \\
                          & \CEO       & \textbf{.808} & \textbf{.880} & \textbf{.876} \\\midrule
\multirow{3}{*}{Combined} & ETypeClus & .396          & .544          & .471          \\
                          & \CEO       & \textbf{.729} & \textbf{.801} & \textbf{.770} \\\bottomrule
\end{tabular}%
}
\caption{Event extraction performance comparison between \CEO and EtypeClus. Recall numbers are recorded to fulfill the goal of extracting as many events as possible. False positives are tolerable since they could be filtered in salient event detection.}
\label{tab:cover_evaluation}
\vspace{-.2in}
\end{table}
We first show the capability of \CEO for \emph{covering more actual event mentions} in \Cref{tab:cover_evaluation}: 1) the transformer model jointly trained for predicate/argument identification and sense disambiguation improves the recall of \textbf{verbal} mentions by around $10\%$ compared with those identified by POS tagging in ETypeClus; 2) with an additional model trained on NomBank for nominal predicates detection, \CEO can capture the majority of \textbf{nominal events} and lead to an overall $30\%$ more events coverage.

\begin{table*}[t!]
\small
\resizebox{\textwidth}{!}{%
\begin{tabular}{@{}lcccccccccc@{}}
\toprule
\multirow{2}{*}{Dataset} & \multicolumn{2}{c}{spkmeans} & \multicolumn{2}{c}{kmeans} & \multicolumn{2}{c}{aggclus} & \multicolumn{2}{c}{jcsc}  & \multicolumn{2}{c}{EtypeClus} \\\cmidrule(lr){2-11}
                         & EtypeClus   & \CEO            & EtypeClus      & \CEO       & EtypeClus       & \CEO       & EtypeClus & \CEO           & EtypeClus   & \CEO             \\\midrule
ACE2005                  & .215        & .350           & .205           & .422      & .157            & .413      & .397      & .525 & .452        & .433            \\
MAVEN                    & .226        & .317           & .199           & .280      & .117            & .367      & .314      & .308          & .326        & .404   \\
RAMS                     & .197        & .246  & .189           & .202      & .186            & .208      & .204      & .214          & .240        & .206  \\ \bottomrule         
\end{tabular}%
}
\caption{Flat clustering performance (ARI) of different algorithms given events represented by EtypeClus and \CEO. Higher scores indicate better performance. Contextualized event embeddings improved by external event knowledge in \CEO help most algorithms achieve much higher ARI than those from EtypeClus. Results evaluated by BCubed-F1 and NMI are similar in \Cref{tab:emb_res_others}.}
\label{tab:emb_res}
\vspace{-.1in}
\end{table*}

Furthermore, we perform flat event clustering with representations learned by \CEO and ETypeClus\footnote{ETypeClus represents events by concatenating predicates and objects, which are not instance-specific but contextual vectors averaged over all occurrences. Conversely, we exclusively represent each event with its respective context considered.}. On the set of common salient events detected by both approaches\footnote{
We find that salient events identified by EtypClus are always covered by \CEO. We therefore directly use salient events identified by ETypeClus. The very few events missed by \CEO can still be represented with sentence embeddings.},
we follow prior work~\cite{shen-etal-2021-corpus} by investigating five clustering algorithms: \emph{kmeans}, Spherical KMeans (\emph{sp-Kmeans}), Agglomerative Clustering(\emph{AggClus}), \emph{JCSC}~\cite{huang-etal-2016-liberal} and \emph{EtypeClus}~\cite{shen-etal-2021-corpus}, and evaluate with three metrics: \emph{ARI}~\cite{hubert1985comparing}, \emph{BCubed-F1}~\cite{bagga-baldwin-1998-entity-based} and \emph{NMI}. We find that results from different metrics are positively related, hence demonstrating performance evaluated by ARI in \Cref{tab:emb_res} and leaving the other two in \Cref{tab:emb_res_others}. In \Cref{tab:emb_res}, we observe significant performance gain when the embeddings learned by \CEO are utilized compared with ETypeClus. We also find that the impact of different event embeddings is less obvious on RAMS, where event types are annotated considering contexts rather than single sentences. 

\begin{table}[t!]
\small
\centering
\resizebox{.85\columnwidth}{!}{%
\begin{tabular}{@{}llccc@{}}\toprule
Event                                                                     & Method    & ACE2005       & MAVEN         & RAMS          \\\midrule
\multirow{4}{*}{\begin{tabular}[c]{@{}l@{}}Mention \\ F1 $\uparrow$\end{tabular}}    & ETypeClus & .132          & .401          & .202          \\
                                                                          & \CEO-NY    & \textbf{.207} & .419          & \textbf{.213} \\
                                                                          & \CEO-DM    & .161          & \textbf{.524} & .199          \\
                                                                          & \CEO-MN    & .141          & .480          & .166          \\\midrule 
\multirow{4}{*}{\begin{tabular}[c]{@{}l@{}}Type \\ Coverage $\uparrow$\end{tabular}} & ETypeClus & .848          & .970          & .885          \\
                                                                          & \CEO-NY    & \textbf{1.0}  & \textbf{1.0}  & \textbf{1.0}  \\
                                                                          & \CEO-DM    & .909          & \textbf{1.0}  & \textbf{1.0}  \\
                                                                          & \CEO-MN    & .909          & \textbf{1.0}  & \textbf{1.0} \\\bottomrule
\end{tabular}%
}
\caption{Performance of event mention detection and type coverage with distant supervision from New York Times (NY), Daily Mail (DM), and Multi-News (MN).}
\label{tab:salient_performance}
\vspace{-.2in}
\end{table}
\paragraph{Benefits of Distant Supervision from Summary Datasets}
We first fine-tune Longformer~\cite{beltagy2020longformer} on three widely-adopted summary datasets for salient event detection: New York Times corpus~\cite{sandhaus2008new}, CNN/Daily Mail~\cite{see-etal-2017-get} and Multi-News~\cite{alex2019multinews}\footnote{For NYT corpus, the events in body texts and their salience labels are provided by~\cite{liu-etal-2018-automatic}. For DailyMail and Multi-News, we extract events triggered by either verbal or nominal predicates with \CEO and automatically annotate them as salient if they also appear in the summary.}. We list salient event detection performance compared with existing approaches on summary datasets in \Cref{tab:summary_salience_performance}. In \Cref{tab:salient_performance}, we show benefits of distant supervision on studied corpora: the model trained on any of the summary datasets is able to capture more salient events compared with ETypeClus, covering all event types. We utilize salient events detected by the model trained on NYT for ontology  and type name generation\footnote{Multiple sources of distant supervision might be helpful for more accurate salient event extraction and we leave this for future work.}.
\begin{figure}[t!]
\centering
\includegraphics[width=\linewidth]{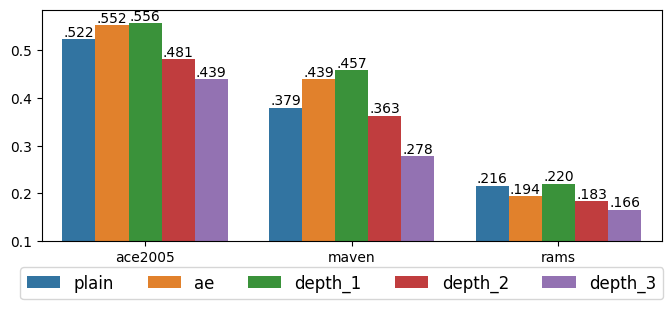}
\caption{Impact of different utilization methods of external WordNet knowledge on hierarchical clustering (\emph{purity} by \emph{linage ward}). When both reconstruction and contrastive loss are employed, we also show the influence of the distance threshold. Dasgupta costs are omitted for statistically insignificant value variances.}
\label{fig:wordnet_performance}
\vspace{-.2in}
\end{figure}
\paragraph{Benefits of External Knowledge on Ontology Inference}
In \Cref{fig:wordnet_performance}, we verify the utility of the external hierarchical event relationship for open-domain ontology induction by comparing performance among \emph{1) \textbf{\textcolor{snsblue}{plain}}}: original embeddings without leveraging external knowledge; \emph{2) \textbf{\textcolor{snsorange}{ae}}}: fine-tuned embeddings only with the reconstruction loss; \emph{3) \textbf{\textcolor{snsgreen}{depth\_1}/\textcolor{snsred}{2}/\textcolor{snspurple}{3}}}: rich embeddings with both reconstruction and contrastive loss. We therefore have the following observations: 1) simply treating event mentions in WordNet as additional instances with the reconstruction loss can hardly guarantee performance gain; 2) selecting event mentions with direct hypernym-hyponym relations (\textcolor{snsgreen}{depth\_1}) as anchors and positives are effective enough to surpass the performance when no external knowledge is utilized.  

\begin{table}[t!]
\resizebox{\linewidth}{!}{%
\begin{tabular}{@{}ll@{}}
\toprule
Topic                     & Event Instances \& Generated Names                                                                                                                        \\ \midrule
\parbox[t]{2mm}{\multirow{12}{*}{\rotatebox[origin=c]{90}{\textbf{Abortion}}}}
& \emph{S1}: Women have to have two in-person doctor appointments prior to \\&receiving an \colorbox{yellow}{\emph{abortion}} and must undergo a state-mandated ultrasound.                     \\
                          & \emph{GPT-J-6B}: \textcolor{snsred}{abortion}                                                                                                                                        \\\cmidrule(l){2-2}
                          & \hypertarget{s2}{\emph{S2}}: ...none would have said "because he will make sure to appoint justices\\ & to the Supreme Court who, given the chance, will \colorbox{yellow}{\emph{overturn}} Roe."                   \\
                          & \emph{GPT-J-6B}: \textcolor{snsred}{abortion:cause:decision:change}                                                                                                                  \\\cmidrule(l){2-2}
                          & \emph{S3}: By a vote of 5-to-4, the court's most conservative members \colorbox{yellow}{\emph{upheld}},\\&for now, a Texas law that, in effect, bans abortions after about six weeks.        \\
                          & \emph{GPT-J-6B}: \textcolor{snsred}{abortion:cause:restrict:app:decision:pass:protect}                                                                                               \\\midrule
\parbox[t]{2mm}{\multirow{12}{*}{\rotatebox[origin=c]{90}{\textbf{LGBT}}}}
                          & \emph{S4}: ...and the First Amendment that the ADF used in the Supreme Court\\& to argue that Phillips shouldn't be required to bake a cake for a  same-sex \\&\colorbox{yellow}{\emph{wedding}}. \\
                          & \emph{GPT-J-6B}: \textcolor{snsred}{make:marriage:wedding}                                                                                                                           \\\cmidrule(l){2-2}
                          & \hypertarget{s5}{\emph{S5}}: The First Amendment Defense Act, as written, would do exactly what\\ & Jeb Bush \colorbox{yellow}{\emph{believes}} -- and much more.                                                \\
                          & \emph{GPT-J-6B}: \textcolor{snsred}{make:change:be:create:think:belief}                                                                                                              \\\cmidrule(l){2-2}
                          & \hypertarget{s6}{\emph{S6}}: ..., 35 percent chose "strongly disapprove," showing passion is higher\\ & among those opposed to marriage \colorbox{yellow}{\emph{equality}}.                                      \\
                          & \emph{GPT-J-6B}: \textcolor{snsred}{make:change:election:cause:equality}                                                                                                             \\ \bottomrule 
\end{tabular}%
}
\caption{Identified events and type names generated by \emph{GPT-J-6B} for instances sampled from two topics. Refer to \Cref{tab:allsides_more_examples_5} and \Cref{tab:allsides_more_examples_4} for the other 9 topics.}
\label{tab:allsides_examples}
\vspace{-.2in}
\end{table}
\subsection{Open-domain Event Ontology Inference}~\label{sec:allsides_exp}
We collect articles over eleven topics from Allsides, including the long-term popular topic \emph{elections} and recently heated debate over \emph{abortion} and \emph{gun control rights}. We consider articles tagged with the same topic as an open domain and show their statistics in \Cref{fig:allsides_statistics}. For events sampled from  \emph{abortion} and \emph{LGBT} corpus, we display the generated type names in \Cref{tab:allsides_examples}, which are highly correlated with their respective topics. 
The finer granularity of names, the more details about events as well as their contexts are reflected. For instance, the event type of the trigger \emph{overturn} (\hyperlink{s2}{\emph{S2}}) is firstly named with the general token \emph{abortion}, then finer token \emph{cause} and \emph{decision}, and lastly the most precise token \emph{change}. We also observe some less appropriate generation, especially among the general type names, such as \emph{make} and \emph{change} for event \emph{believes} (\hyperlink{s5}{\emph{S5}}) and \emph{equality} (\hyperlink{s6}{\emph{S6}}). We attribute the less accurate coarse types to the single root restriction for the induced event ontology and leave multi-root ontology induction for future investigation.

\section{Conclusion}
To understand events expressed in open domains free from the restriction of pre-defined ontologies, we propose a new Corpus-based open-domain Event Ontology induction strategy \CEO to automatically induce hierarchical event ontology structure and provide interpretable type names for further curation. On three  event datasets, we find it can capture salient events more accurately, induce ontology structures aligning well with ground truth and generate appropriate coarse-to-fine type names. We also show the broad application of \CEO on  open domains from Allsides. 

\section*{Limitations}
An important caveat to this work is the assumption that all event types in the studied open-domain corpus could be covered by a single tree-structured schema. However, sometimes events in a corpus could be quite different and we can hardly categorize them with a single coarse type as the root node of the ontology tree. Meanwhile, we restrict the induced event ontology in a tree structure. Although event schemas pre-defined by humans in popular event datasets follow the tree structure, it is likely other styles of ontology can better describe events and their relations in emerging corpora. As the first event ontology induction model that can induce a hierarchical event ontology with meaningful names, we advocate more efforts in exploring event ontology in the open-domain setting. 
\section*{Ethical Consideration}
\CEO is an effective strategy for event ontology induction that leverages widely-adopted textual data and NLP models pretrained on fairly neutral corpora. To the best of our knowledge, \CEO helps understand events from all studied datasets in this paper without raising privacy issues or increasing bias in the induced event ontology. 
\bibliography{anthology,custom}

\begin{thebibliography}{54}
\expandafter\ifx\csname natexlab\endcsname\relax\def\natexlab#1{#1}\fi

\bibitem[{Bagga and Baldwin(1998)}]{bagga-baldwin-1998-entity-based}
Amit Bagga and Breck Baldwin. 1998.
\newblock \href {https://doi.org/10.3115/980845.980859} {Entity-based
  cross-document coreferencing using the vector space model}.
\newblock In \emph{36th Annual Meeting of the Association for Computational
  Linguistics and 17th International Conference on Computational Linguistics,
  Volume 1}, pages 79--85, Montreal, Quebec, Canada. Association for
  Computational Linguistics.

\bibitem[{Baker et~al.(1998)Baker, Fillmore, and Lowe}]{baker1998berkeley}
Collin~F Baker, Charles~J Fillmore, and John~B Lowe. 1998.
\newblock The berkeley framenet project.
\newblock In \emph{COLING 1998 Volume 1: The 17th International Conference on
  Computational Linguistics}.

\bibitem[{Balntas et~al.(2016)Balntas, Riba, Ponsa, and
  Mikolajczyk}]{balntas2016learning}
Vassileios Balntas, Edgar Riba, Daniel Ponsa, and Krystian Mikolajczyk. 2016.
\newblock Learning local feature descriptors with triplets and shallow
  convolutional neural networks.
\newblock In \emph{Bmvc}, volume~1, page~3.

\bibitem[{Beltagy et~al.(2020)Beltagy, Peters, and
  Cohan}]{beltagy2020longformer}
Iz~Beltagy, Matthew~E Peters, and Arman Cohan. 2020.
\newblock Longformer: The long-document transformer.
\newblock \emph{arXiv preprint arXiv:2004.05150}.

\bibitem[{Bougouin et~al.(2013)Bougouin, Boudin, and
  Daille}]{bougouin-etal-2013-topicrank}
Adrien Bougouin, Florian Boudin, and B{\'e}atrice Daille. 2013.
\newblock \href {https://aclanthology.org/I13-1062} {{T}opic{R}ank: Graph-based
  topic ranking for keyphrase extraction}.
\newblock In \emph{Proceedings of the Sixth International Joint Conference on
  Natural Language Processing}, pages 543--551, Nagoya, Japan. Asian Federation
  of Natural Language Processing.

\bibitem[{Brown et~al.(2020)Brown, Mann, Ryder, Subbiah, Kaplan, Dhariwal,
  Neelakantan, Shyam, Sastry, Askell et~al.}]{brown2020language}
Tom Brown, Benjamin Mann, Nick Ryder, Melanie Subbiah, Jared~D Kaplan, Prafulla
  Dhariwal, Arvind Neelakantan, Pranav Shyam, Girish Sastry, Amanda Askell,
  et~al. 2020.
\newblock Language models are few-shot learners.
\newblock \emph{Advances in neural information processing systems},
  33:1877--1901.

\bibitem[{Chami et~al.(2020)Chami, Gu, Chatziafratis, and
  R{\'e}}]{chami2020trees}
Ines Chami, Albert Gu, Vaggos Chatziafratis, and Christopher R{\'e}. 2020.
\newblock From trees to continuous embeddings and back: Hyperbolic hierarchical
  clustering.
\newblock \emph{Advances in Neural Information Processing Systems},
  33:15065--15076.

\bibitem[{Chen et~al.(2021)Chen, Zhang, Ning, Li, Ji, McKeown, and
  Roth}]{chen-etal-2021-event}
Muhao Chen, Hongming Zhang, Qiang Ning, Manling Li, Heng Ji, Kathleen McKeown,
  and Dan Roth. 2021.
\newblock \href {https://doi.org/10.18653/v1/2021.acl-tutorials.2}
  {Event-centric natural language processing}.
\newblock In \emph{Proceedings of the 59th Annual Meeting of the Association
  for Computational Linguistics and the 11th International Joint Conference on
  Natural Language Processing: Tutorial Abstracts}, pages 6--14, Online.
  Association for Computational Linguistics.

\bibitem[{Dasgupta(2016)}]{dasgupta2016cost}
Sanjoy Dasgupta. 2016.
\newblock A cost function for similarity-based hierarchical clustering.
\newblock In \emph{Proceedings of the forty-eighth annual ACM symposium on
  Theory of Computing}, pages 118--127.

\bibitem[{Devlin et~al.(2019)Devlin, Chang, Lee, and
  Toutanova}]{devlin-etal-2019-bert}
Jacob Devlin, Ming-Wei Chang, Kenton Lee, and Kristina Toutanova. 2019.
\newblock \href {https://doi.org/10.18653/v1/N19-1423} {{BERT}: Pre-training of
  deep bidirectional transformers for language understanding}.
\newblock In \emph{Proceedings of the 2019 Conference of the North {A}merican
  Chapter of the Association for Computational Linguistics: Human Language
  Technologies, Volume 1 (Long and Short Papers)}, pages 4171--4186,
  Minneapolis, Minnesota. Association for Computational Linguistics.

\bibitem[{Doddington et~al.(2004)Doddington, Mitchell, Przybocki, Ramshaw,
  Strassel, and Weischedel}]{doddington-etal-2004-automatic}
George Doddington, Alexis Mitchell, Mark Przybocki, Lance Ramshaw, Stephanie
  Strassel, and Ralph Weischedel. 2004.
\newblock \href {http://www.lrec-conf.org/proceedings/lrec2004/pdf/5.pdf} {The
  automatic content extraction ({ACE}) program {--} tasks, data, and
  evaluation}.
\newblock In \emph{Proceedings of the Fourth International Conference on
  Language Resources and Evaluation ({LREC}{'}04)}, Lisbon, Portugal. European
  Language Resources Association (ELRA).

\bibitem[{Domingos(2015)}]{domingos2015master}
Pedro Domingos. 2015.
\newblock \emph{The master algorithm: How the quest for the ultimate learning
  machine will remake our world}.
\newblock Basic Books.

\bibitem[{Ebner et~al.(2020)Ebner, Xia, Culkin, Rawlins, and
  Van~Durme}]{ebner-etal-2020-multi}
Seth Ebner, Patrick Xia, Ryan Culkin, Kyle Rawlins, and Benjamin Van~Durme.
  2020.
\newblock \href {https://doi.org/10.18653/v1/2020.acl-main.718} {Multi-sentence
  argument linking}.
\newblock In \emph{Proceedings of the 58th Annual Meeting of the Association
  for Computational Linguistics}, pages 8057--8077, Online. Association for
  Computational Linguistics.

\bibitem[{Edwards and Ji(2022)}]{edwards2022semi}
Carl Edwards and Heng Ji. 2022.
\newblock Semi-supervised new event type induction and description via
  contrastive loss-enforced batch attention.
\newblock \emph{arXiv preprint arXiv:2202.05943}.

\bibitem[{Fabbri et~al.(2019)Fabbri, Li, She, Li, and
  Radev}]{alex2019multinews}
Alexander~R. Fabbri, Irene Li, Tianwei She, Suyi Li, and Dragomir~R. Radev.
  2019.
\newblock \href {http://arxiv.org/abs/1906.01749} {Multi-news: a large-scale
  multi-document summarization dataset and abstractive hierarchical model}.

\bibitem[{Fellbaum(2010)}]{fellbaum2010wordnet}
Christiane Fellbaum. 2010.
\newblock Wordnet.
\newblock In \emph{Theory and applications of ontology: computer applications},
  pages 231--243. Springer.

\bibitem[{Fung et~al.(2021)Fung, Thomas, Gangi~Reddy, Polisetty, Ji, Chang,
  McKeown, Bansal, and Sil}]{fung-etal-2021-infosurgeon}
Yi~Fung, Christopher Thomas, Revanth Gangi~Reddy, Sandeep Polisetty, Heng Ji,
  Shih-Fu Chang, Kathleen McKeown, Mohit Bansal, and Avi Sil. 2021.
\newblock \href {https://doi.org/10.18653/v1/2021.acl-long.133}
  {{I}nfo{S}urgeon: Cross-media fine-grained information consistency checking
  for fake news detection}.
\newblock In \emph{Proceedings of the 59th Annual Meeting of the Association
  for Computational Linguistics and the 11th International Joint Conference on
  Natural Language Processing (Volume 1: Long Papers)}, pages 1683--1698,
  Online. Association for Computational Linguistics.

\bibitem[{Garofolo et~al.(1993)Garofolo, Graff, Paul, and
  Pallett}]{garofolo1993csr}
John Garofolo, David Graff, Doug Paul, and David Pallett. 1993.
\newblock Csr-i (wsj0) complete ldc93s6a.
\newblock \emph{Web Download. Philadelphia: Linguistic Data Consortium}, 83.

\bibitem[{Grootendorst(2020)}]{grootendorst2020keybert}
Maarten Grootendorst. 2020.
\newblock \href {https://doi.org/10.5281/zenodo.4461265} {Keybert: Minimal
  keyword extraction with bert.}

\bibitem[{Guzman-Nateras et~al.(2022)Guzman-Nateras, Nguyen, and
  Nguyen}]{guzman-nateras-etal-2022-cross}
Luis Guzman-Nateras, Minh~Van Nguyen, and Thien Nguyen. 2022.
\newblock \href {https://doi.org/10.18653/v1/2022.naacl-main.409}
  {Cross-lingual event detection via optimized adversarial training}.
\newblock In \emph{Proceedings of the 2022 Conference of the North American
  Chapter of the Association for Computational Linguistics: Human Language
  Technologies}, pages 5588--5599, Seattle, United States. Association for
  Computational Linguistics.

\bibitem[{Heller and Ghahramani(2005)}]{heller2005bayesian}
Katherine~A Heller and Zoubin Ghahramani. 2005.
\newblock Bayesian hierarchical clustering.
\newblock In \emph{Proceedings of the 22nd international conference on Machine
  learning}, pages 297--304.

\bibitem[{Hsu et~al.(2022)Hsu, Huang, Boschee, Miller, Natarajan, Chang, and
  Peng}]{hsu-etal-2022-degree}
I-Hung Hsu, Kuan-Hao Huang, Elizabeth Boschee, Scott Miller, Prem Natarajan,
  Kai-Wei Chang, and Nanyun Peng. 2022.
\newblock \href {https://doi.org/10.18653/v1/2022.naacl-main.138} {{DEGREE}: A
  data-efficient generation-based event extraction model}.
\newblock In \emph{Proceedings of the 2022 Conference of the North American
  Chapter of the Association for Computational Linguistics: Human Language
  Technologies}, pages 1890--1908, Seattle, United States. Association for
  Computational Linguistics.

\bibitem[{Huang et~al.(2016)Huang, Cassidy, Feng, Ji, Voss, Han, and
  Sil}]{huang-etal-2016-liberal}
Lifu Huang, Taylor Cassidy, Xiaocheng Feng, Heng Ji, Clare~R. Voss, Jiawei Han,
  and Avirup Sil. 2016.
\newblock \href {https://doi.org/10.18653/v1/P16-1025} {Liberal event
  extraction and event schema induction}.
\newblock In \emph{Proceedings of the 54th Annual Meeting of the Association
  for Computational Linguistics (Volume 1: Long Papers)}, pages 258--268,
  Berlin, Germany. Association for Computational Linguistics.

\bibitem[{Huang and Ji(2020)}]{huang-ji-2020-semi}
Lifu Huang and Heng Ji. 2020.
\newblock \href {https://doi.org/10.18653/v1/2020.emnlp-main.53}
  {Semi-supervised new event type induction and event detection}.
\newblock In \emph{Proceedings of the 2020 Conference on Empirical Methods in
  Natural Language Processing (EMNLP)}, pages 718--724, Online. Association for
  Computational Linguistics.

\bibitem[{Hubert and Arabie(1985)}]{hubert1985comparing}
Lawrence Hubert and Phipps Arabie. 1985.
\newblock Comparing partitions.
\newblock \emph{Journal of classification}, 2(1):193--218.

\bibitem[{Jindal et~al.(2020)Jindal, Deutsch, and
  Roth}]{jindal-etal-2020-killed}
Disha Jindal, Daniel Deutsch, and Dan Roth. 2020.
\newblock \href {https://doi.org/10.18653/v1/2020.coling-main.10} {Is killed
  more significant than fled? a contextual model for salient event detection}.
\newblock In \emph{Proceedings of the 28th International Conference on
  Computational Linguistics}, pages 114--124, Barcelona, Spain (Online).
  International Committee on Computational Linguistics.

\bibitem[{Kobren et~al.(2017)Kobren, Monath, Krishnamurthy, and
  McCallum}]{kobren2017hierarchical}
Ari Kobren, Nicholas Monath, Akshay Krishnamurthy, and Andrew McCallum. 2017.
\newblock A hierarchical algorithm for extreme clustering.
\newblock In \emph{Proceedings of the 23rd ACM SIGKDD international conference
  on knowledge discovery and data mining}, pages 255--264.

\bibitem[{Lee et~al.(2021)Lee, Tiha, Yuqian, and Hegler}]{SRLEnglish2021}
Celine Lee, Anjana Tiha, Deng Yuqian, and Tissot Hegler. 2021.
\newblock English semantic role labeling (srl) demo.
\newblock \url{https://github.com/CogComp/SRL-English}.

\bibitem[{Lin(2004)}]{lin-2004-rouge}
Chin-Yew Lin. 2004.
\newblock \href {https://aclanthology.org/W04-1013} {{ROUGE}: A package for
  automatic evaluation of summaries}.
\newblock In \emph{Text Summarization Branches Out}, pages 74--81, Barcelona,
  Spain. Association for Computational Linguistics.

\bibitem[{Lin et~al.(2020)Lin, Ji, Huang, and Wu}]{lin-etal-2020-joint}
Ying Lin, Heng Ji, Fei Huang, and Lingfei Wu. 2020.
\newblock \href {https://doi.org/10.18653/v1/2020.acl-main.713} {A joint neural
  model for information extraction with global features}.
\newblock In \emph{Proceedings of the 58th Annual Meeting of the Association
  for Computational Linguistics}, pages 7999--8009, Online. Association for
  Computational Linguistics.

\bibitem[{Liu et~al.(2019)Liu, Ott, Goyal, Du, Joshi, Chen, Levy, Lewis,
  Zettlemoyer, and Stoyanov}]{liu2019roberta}
Yinhan Liu, Myle Ott, Naman Goyal, Jingfei Du, Mandar Joshi, Danqi Chen, Omer
  Levy, Mike Lewis, Luke Zettlemoyer, and Veselin Stoyanov. 2019.
\newblock Roberta: A robustly optimized bert pretraining approach.
\newblock \emph{arXiv preprint arXiv:1907.11692}.

\bibitem[{Liu et~al.(2018)Liu, Xiong, Mitamura, and
  Hovy}]{liu-etal-2018-automatic}
Zhengzhong Liu, Chenyan Xiong, Teruko Mitamura, and Eduard Hovy. 2018.
\newblock \href {https://doi.org/10.18653/v1/D18-1154} {Automatic event
  salience identification}.
\newblock In \emph{Proceedings of the 2018 Conference on Empirical Methods in
  Natural Language Processing}, pages 1226--1236, Brussels, Belgium.
  Association for Computational Linguistics.

\bibitem[{Lu et~al.(2022)Lu, Hsu, Zhou, Ma, and
  Chen}]{lu-etal-2022-summarization}
Keming Lu, I-Hung Hsu, Wenxuan Zhou, Mingyu~Derek Ma, and Muhao Chen. 2022.
\newblock \href {https://aclanthology.org/2022.findings-emnlp.490}
  {Summarization as indirect supervision for relation extraction}.
\newblock In \emph{Findings of the Association for Computational Linguistics:
  EMNLP 2022}, pages 6575--6594, Abu Dhabi, United Arab Emirates. Association
  for Computational Linguistics.

\bibitem[{Meyers et~al.(2004)Meyers, Reeves, Macleod, Szekely, Zielinska,
  Young, and Grishman}]{meyers-etal-2004-nombank}
Adam Meyers, Ruth Reeves, Catherine Macleod, Rachel Szekely, Veronika
  Zielinska, Brian Young, and Ralph Grishman. 2004.
\newblock \href {https://aclanthology.org/W04-2705} {The {N}om{B}ank project:
  An interim report}.
\newblock In \emph{Proceedings of the Workshop Frontiers in Corpus Annotation
  at {HLT}-{NAACL} 2004}, pages 24--31, Boston, Massachusetts, USA. Association
  for Computational Linguistics.

\bibitem[{Mihalcea and Tarau(2004)}]{mihalcea-tarau-2004-textrank}
Rada Mihalcea and Paul Tarau. 2004.
\newblock \href {https://aclanthology.org/W04-3252} {{T}ext{R}ank: Bringing
  order into text}.
\newblock In \emph{Proceedings of the 2004 Conference on Empirical Methods in
  Natural Language Processing}, pages 404--411, Barcelona, Spain. Association
  for Computational Linguistics.

\bibitem[{Monath et~al.(2019)Monath, Zaheer, Silva, McCallum, and
  Ahmed}]{monath2019gradient}
Nicholas Monath, Manzil Zaheer, Daniel Silva, Andrew McCallum, and Amr Ahmed.
  2019.
\newblock Gradient-based hierarchical clustering using continuous
  representations of trees in hyperbolic space.
\newblock In \emph{Proceedings of the 25th ACM SIGKDD International Conference
  on Knowledge Discovery \& Data Mining}, pages 714--722.

\bibitem[{Moseley and Wang(2017)}]{moseley2017approximation}
Benjamin Moseley and Joshua Wang. 2017.
\newblock Approximation bounds for hierarchical clustering: Average linkage,
  bisecting k-means, and local search.
\newblock \emph{Advances in neural information processing systems}, 30.

\bibitem[{Pradhan et~al.(2013)Pradhan, Moschitti, Xue, Ng, Bj{\"o}rkelund,
  Uryupina, Zhang, and Zhong}]{pradhan2013towards}
Sameer Pradhan, Alessandro Moschitti, Nianwen Xue, Hwee~Tou Ng, Anders
  Bj{\"o}rkelund, Olga Uryupina, Yuchen Zhang, and Zhi Zhong. 2013.
\newblock Towards robust linguistic analysis using ontonotes.
\newblock In \emph{Proceedings of the Seventeenth Conference on Computational
  Natural Language Learning}, pages 143--152.

\bibitem[{Reimers and
  Gurevych(2019{\natexlab{a}})}]{reimers-gurevych-2019-sentence}
Nils Reimers and Iryna Gurevych. 2019{\natexlab{a}}.
\newblock \href {https://doi.org/10.18653/v1/D19-1410} {Sentence-{BERT}:
  Sentence embeddings using {S}iamese {BERT}-networks}.
\newblock In \emph{Proceedings of the 2019 Conference on Empirical Methods in
  Natural Language Processing and the 9th International Joint Conference on
  Natural Language Processing (EMNLP-IJCNLP)}, pages 3982--3992, Hong Kong,
  China. Association for Computational Linguistics.

\bibitem[{Reimers and Gurevych(2019{\natexlab{b}})}]{reimers2019sentence}
Nils Reimers and Iryna Gurevych. 2019{\natexlab{b}}.
\newblock Sentence-bert: Sentence embeddings using siamese bert-networks.
\newblock \emph{arXiv preprint arXiv:1908.10084}.

\bibitem[{Sandhaus(2008)}]{sandhaus2008new}
Evan Sandhaus. 2008.
\newblock The new york times annotated corpus.
\newblock \emph{Linguistic Data Consortium, Philadelphia}, 6(12):e26752.

\bibitem[{See et~al.(2017)See, Liu, and Manning}]{see-etal-2017-get}
Abigail See, Peter~J. Liu, and Christopher~D. Manning. 2017.
\newblock \href {https://doi.org/10.18653/v1/P17-1099} {Get to the point:
  Summarization with pointer-generator networks}.
\newblock In \emph{Proceedings of the 55th Annual Meeting of the Association
  for Computational Linguistics (Volume 1: Long Papers)}, pages 1073--1083,
  Vancouver, Canada. Association for Computational Linguistics.

\bibitem[{Shen et~al.(2021)Shen, Zhang, Ji, and Han}]{shen-etal-2021-corpus}
Jiaming Shen, Yunyi Zhang, Heng Ji, and Jiawei Han. 2021.
\newblock \href {https://doi.org/10.18653/v1/2021.emnlp-main.441} {Corpus-based
  open-domain event type induction}.
\newblock In \emph{Proceedings of the 2021 Conference on Empirical Methods in
  Natural Language Processing}, pages 5427--5440, Online and Punta Cana,
  Dominican Republic. Association for Computational Linguistics.

\bibitem[{Shi and Lin(2019)}]{shi2019simple}
Peng Shi and Jimmy Lin. 2019.
\newblock Simple bert models for relation extraction and semantic role
  labeling.
\newblock \emph{arXiv preprint arXiv:1904.05255}.

\bibitem[{Wang and Komatsuzaki(2021)}]{wang2021gpt}
Ben Wang and Aran Komatsuzaki. 2021.
\newblock Gpt-j-6b: A 6 billion parameter autoregressive language model.

\bibitem[{Wang and Wang(2018)}]{wang2018improved}
Dingkang Wang and Yusu Wang. 2018.
\newblock An improved cost function for hierarchical cluster trees.
\newblock \emph{arXiv preprint arXiv:1812.02715}.

\bibitem[{Wang et~al.(2022)Wang, Yu, Chang, Sun, and
  Huang}]{wang-etal-2022-query}
Sijia Wang, Mo~Yu, Shiyu Chang, Lichao Sun, and Lifu Huang. 2022.
\newblock \href {https://doi.org/10.18653/v1/2022.findings-acl.16} {Query and
  extract: Refining event extraction as type-oriented binary decoding}.
\newblock In \emph{Findings of the Association for Computational Linguistics:
  ACL 2022}, pages 169--182, Dublin, Ireland. Association for Computational
  Linguistics.

\bibitem[{Wang et~al.(2020)Wang, Wang, Han, Jiang, Han, Liu, Li, Li, Lin, and
  Zhou}]{wang-etal-2020-maven}
Xiaozhi Wang, Ziqi Wang, Xu~Han, Wangyi Jiang, Rong Han, Zhiyuan Liu, Juanzi
  Li, Peng Li, Yankai Lin, and Jie Zhou. 2020.
\newblock \href {https://doi.org/10.18653/v1/2020.emnlp-main.129} {{MAVEN}: {A}
  {M}assive {G}eneral {D}omain {E}vent {D}etection {D}ataset}.
\newblock In \emph{Proceedings of the 2020 Conference on Empirical Methods in
  Natural Language Processing (EMNLP)}, pages 1652--1671, Online. Association
  for Computational Linguistics.

\bibitem[{Ward~Jr(1963)}]{ward1963hierarchical}
Joe~H Ward~Jr. 1963.
\newblock Hierarchical grouping to optimize an objective function.
\newblock \emph{Journal of the American statistical association},
  58(301):236--244.

\bibitem[{Zhang et~al.(2022)Zhang, Ji, Ji, and Wang}]{zhang-etal-2022-zero}
Senhui Zhang, Tao Ji, Wendi Ji, and Xiaoling Wang. 2022.
\newblock \href {https://doi.org/10.18653/v1/2022.findings-naacl.196}
  {Zero-shot event detection based on ordered contrastive learning and
  prompt-based prediction}.
\newblock In \emph{Findings of the Association for Computational Linguistics:
  NAACL 2022}, pages 2572--2580, Seattle, United States. Association for
  Computational Linguistics.

\bibitem[{Zhang et~al.(1997)Zhang, Ramakrishnan, and Livny}]{zhang1997birch}
Tian Zhang, Raghu Ramakrishnan, and Miron Livny. 1997.
\newblock Birch: A new data clustering algorithm and its applications.
\newblock \emph{Data mining and knowledge discovery}, 1(2):141--182.

\bibitem[{Zhang et~al.(2020)Zhang, Chen, and Bui}]{zhang2020diagnostic}
Tianran Zhang, Muhao Chen, and Alex~AT Bui. 2020.
\newblock Diagnostic prediction with sequence-of-sets representation learning
  for clinical events.
\newblock In \emph{International Conference on Artificial Intelligence in
  Medicine}, pages 348--358. Springer.

\bibitem[{Zhang et~al.(2019)Zhang, Kishore, Wu, Weinberger, and
  Artzi}]{zhang2019bertscore}
Tianyi Zhang, Varsha Kishore, Felix Wu, Kilian~Q Weinberger, and Yoav Artzi.
  2019.
\newblock Bertscore: Evaluating text generation with bert.
\newblock \emph{arXiv preprint arXiv:1904.09675}.

\bibitem[{Zhang et~al.(2023)Zhang, Elfardy, Dreyer, Small, Ji, and
  Bansal}]{zhang-etal-2023-enhancing}
Zixuan Zhang, Heba Elfardy, Markus Dreyer, Kevin Small, Heng Ji, and Mohit
  Bansal. 2023.
\newblock \href {https://aclanthology.org/2023.eacl-main.124} {Enhancing
  multi-document summarization with cross-document graph-based information
  extraction}.
\newblock In \emph{Proceedings of the 17th Conference of the European Chapter
  of the Association for Computational Linguistics}, pages 1696--1707,
  Dubrovnik, Croatia. Association for Computational Linguistics.

\end{thebibliography}
\bibliographystyle{acl_natbib}
\clearpage
\appendix
\section{Appendix}\label{sec:appendix}
\subsection{Evaluation Metrics}
\paragraph{Hierarchical Clustering} \label{sec:hierarchical_clustering_metric}
As discussed in \Cref{sec:hierarchical_clustering}, we leverage the following two metrics to compare the induced event ontologies with the ground truth:
\begin{itemize}
\item \emph{Dendrogram Purity}~\cite{heller2005bayesian}: Given the dataset $X$, the $k$-th ground-truth flat cluster $\mathcal{C}_k^*$ and the inferred tree structure $\mathcal{T}$, dendrogram purity is the average purity of the least common ancestors of pairs of points belonging to the same ground truth cluster:
\small
\[
P(\mathcal{T})=\frac{1}{|\mathcal{P}^*|}\sum_{k=1}^K\sum_{x_i,x_j\in\mathcal{C}_k^*}\text{pur}\Big(\underbrace{\text{lvs}\big(\text{lca}(x_i,x_j)\big)}_{\text{inferred }\mathcal{T}}, \mathcal{C}_k^*\Big),
\]
\normalsize
where $|\mathcal{P}^*|$ represents the number of data point pairs in the same ground-truth cluster, $\text{lca}(x_i, x_j)$ gives the least common ancestor of $x_i$ and $x_j$ in the inferred tree $\mathcal{T}$, $\text{lvs} (n)$ gives a set of leaf node descendants of node $n$, while $\text{pur}(\cdot,\cdot)$ measures the fraction of data points under its first cluster (i.e., the inferred cluster) that are members of the second (i.e., the ground-truth cluster).
\item \emph{Dasgupta's Cost}~\cite{dasgupta2016cost}: Good trees acknowledged by Dasgupta cost should cluster data such that similar data points have least common ancestors much further from the root than that of dissimilar data points:
\[
C(\mathcal{T})=\sum_{x_i,x_j\in X}\omega_{i,j}|\text{lvs}\big(\text{lca}(x_i,x_j)\big)|,
\]
where $\omega_{i,j}$ measures pairwise similarity. In summary, inferred trees with higher purity and lower cost achieve more accurate hierarchical event clustering.
\end{itemize}
\paragraph{Name Generation}\label{sec:name_generation_metrics}
\emph{Sim dist} is self-defined to consider both semantic similarity and granularity difference between each pair of reference $e^i_r$ and generated name $e^j_p$:
\begin{align}
sim\_dist&=1/(n_r\cdot n_p)\sum_{i,j}\underbrace{\big(1-|i/n_r-j/n_p|\big)}_{\text{granularity difference}}\cdot\nonumber\\
&\underbrace{\Big(\cos\big(emb(e^i_r),emb(e^j_p)\big)+1\Big)/2}_{\text{semantic similarity}},
\nonumber
\end{align}
where $emb$ is phrase representation from SBERT~\cite{reimers2019sentence}.
\subsection{Baselines}
\paragraph{Hierarchical Clustering}
\begin{itemize}
\item \emph{Hierarchical Kmeans}: it splits data into two clusters at each iteration using Kmeans~\footnote{We use Bisecting K-Means as the direct analog of hierarchical KMeans~\cite{moseley2017approximation}.}.
\item \emph{Birch}~\cite{zhang1997birch}: it adopts a dynamically growing tree structure with points inserted greedily using the node statistics and split operation invoked when the branching factor is exceeded.
\item \emph{Ward Linkage}~\cite{ward1963hierarchical}: the algorithm uses the Ward variance minimization algorithm to calculate the distance between the newly formed cluster and other clusters in the forest.
\item \emph{Perch}~\cite{kobren2017hierarchical}: it incrementally builds a tree structure by inserting points as a sibling of their nearest neighbor and performs local tree re-arrangements.
\item \emph{gHHC}~\cite{monath2019gradient}: it represents uncertainty over tree structures with vectors in the Poincaré ball and optimizes hyperbolic embeddings of internal nodes using an objective related to Dasgupta's cost~\cite{dasgupta2016cost,wang2018improved}.
\item \emph{HypHC}~\cite{chami2020trees}: it derives a continuous relaxation of Dasgupta's discrete objective~\cite{dasgupta2016cost} by introducing a continuous analog for the notion of the lowest common ancestor.
\end{itemize}
\paragraph{Name Generation}\label{sec:name_generation_strategy}
\begin{itemize}
\item \emph{Most frequent}: the token that appears most in the event triggers are extracted as the cluster name. 
\item \emph{tf-idf}: following~\cite{shen-etal-2021-corpus}, we obtain more popular trigger tokens in the studied corpus with regard to their frequency in general corpora. 
\item \emph{textrank}~\cite{mihalcea-tarau-2004-textrank}, \emph{topicrank}~\cite{bougouin-etal-2013-topicrank} and \emph{KeyBERT}~\cite{grootendorst2020keybert}: we cast the cluster name generation as the keyword extraction task, hence the above three strategies are utilized to extract keywords given sentences from the same cluster. 
\item \emph{wordnet synset}: since WordNet~\cite{fellbaum2010wordnet} describes the relatedness of word synsets in the hypernym-hyponym format, we introduce the \emph{wordnet synset} strategy where the cluster is named after the least common ancestor hypernym of event triggers. 
\item \emph{RoBERTa}~\cite{liu2019roberta}: given the context of even triggers, the masked language model \emph{RoBERTa-large} is employed to obtain token probabilities of the trigger position and the token with the highest probability over all instances is adopted as the cluster name. 
\item\emph{GPT-J}~\cite{wang2021gpt}: motivated by the in-context learning capabilities of generative language models~\cite{brown2020language}, we provide the sentence, the trigger phrase as well as the finest label name of instances sampled from other corpora as the demonstration and acquire the label distribution of testing instances from \emph{GPT-J-6B}~\footnote{In the unsupervised setting, we use examples from other datasets to provide the finest label name required in the demonstrations. Similar to RoBERTa, the output token with the highest probability across instances in the same cluster is adopted as the label name.}.
\end{itemize}
\subsection{Autoencoder Design to Improve Event Embeddings}\label{sec:autoencoder_design}
\begin{figure*}[t!]
\centering
\includegraphics[width=\textwidth]{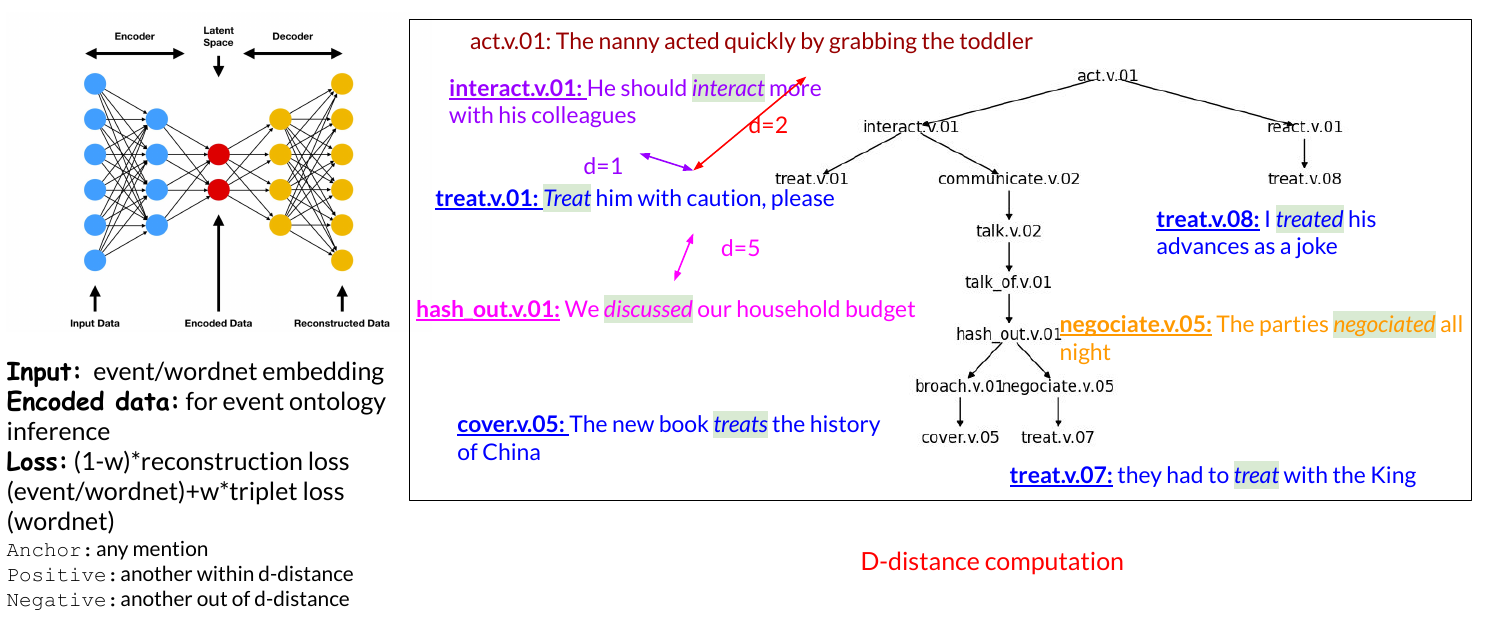}
\caption{The proposed autoencoder model to improve event embeddings by leveraging external knowledge. The typical autoencoder architecture is optimized with the weighted sum of reconstruction loss and contrastive triplet margin loss (left). The event mention triplet in the form of <anchor, positive, negative> is selected based on the $d$-distance, which is calculated according to the pre-defined ontology of WordNet (right). }
\label{fig:external_wordnet}
\end{figure*}
As introduced in \Cref{sec:method_ontology_inference}, an autoencoder optimized by reconstruction and triplet loss exploits external event knowledge from WordNet. To extract anchor synsets and their corresponding positive and negative ones, we first define the distance between different synsets in the ontology tree.  Considering the synset \emph{treat.v.01} in the partial ontology demonstrated in \Cref{fig:external_wordnet} as an anchor event: its distance to the first-level hypernym \emph{interact.v.01} is 1 and the second-level hypernym \emph{act.v.01} is 2; furthermore, its distance to the loosely related synset \emph{hash\_out.v.01} is $5$. Suppose the threshold distance to distinguish positive from negative events is 2, then we treat \emph{interact.v.01} and \emph{act.v.01} as positive event mentions while \emph{hash\_out.v.01} as the negative. 
\begin{table}[h!]
\resizebox{\columnwidth}{!}{%
\begin{tabular}{@{}lll@{}}
\toprule
\multicolumn{2}{l}{Template}         & Demonstration                                                                                                                            \\ \midrule
\multirow{2}{*}{Input} & sentence:   & \textit{\begin{tabular}[c]{@{}l@{}}Do you think Arafat's death will help or \\ hurt the Israeli-Palestinian peace process?\end{tabular}} \\\cmidrule(l){2-3} 
                       & predicate:  & \textit{death}                                                                                                                           \\\midrule
Output                 & event type: & \textit{Die}                                                                                                                             \\ \bottomrule
\end{tabular}%
}
\caption{Example input-output pair for event type name generation. To retrieve the event type of a test instance, several demonstrations with input and output are randomly sampled and the token with the maximum probability from the PLM is adopted as the type name.}
\label{tab:name_generation_prompt}
\end{table}

\begin{figure}[h!]
     \centering
     \begin{subfigure}[b]{\linewidth}
         \centering
         \includegraphics[width=\linewidth]{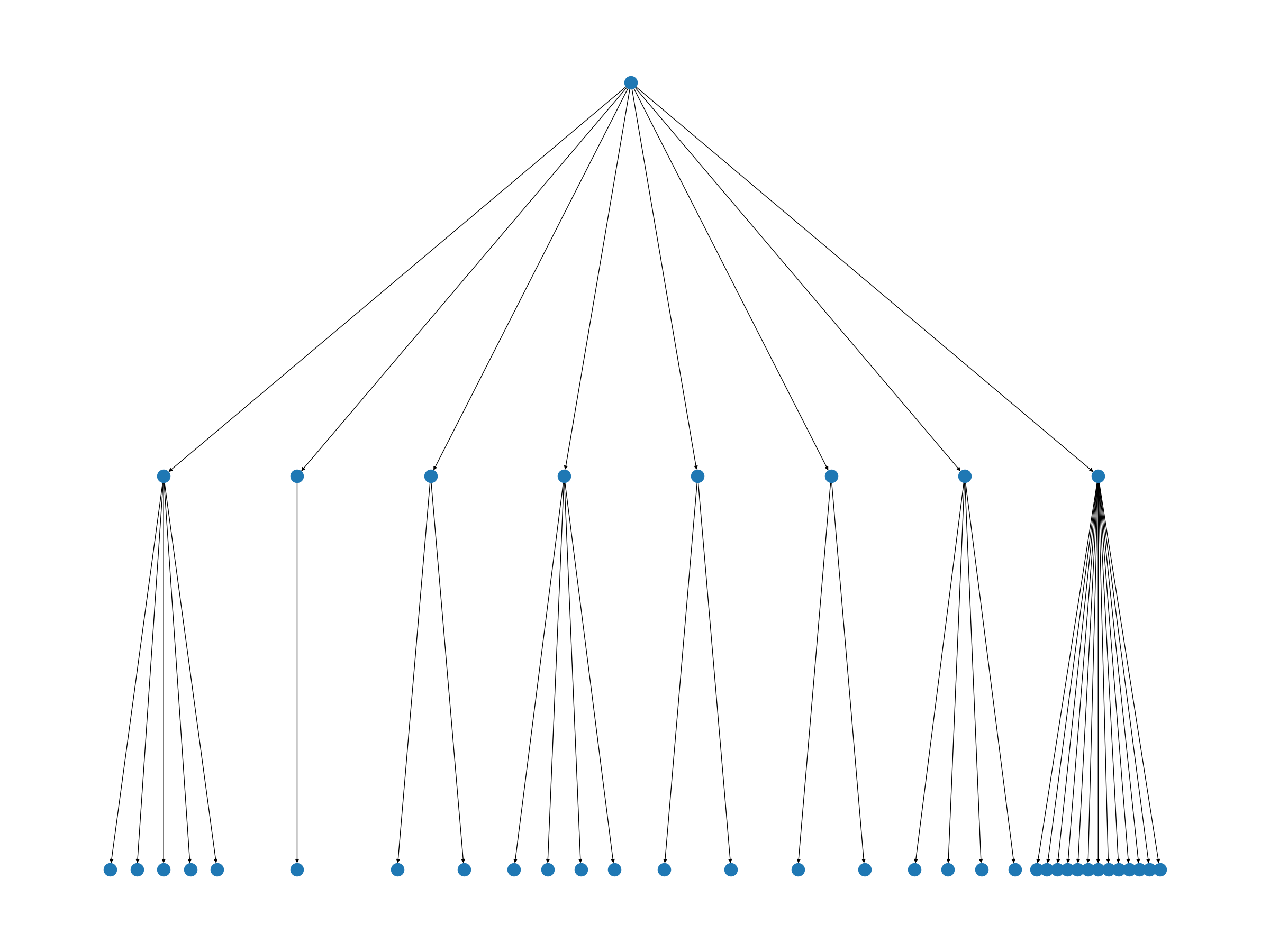}
         \caption{ACE 2005}
         \label{fig:ace2005_ontology}
     \end{subfigure}
     \hfill
     \begin{subfigure}[b]{\linewidth}
         \centering
         \includegraphics[width=\linewidth]{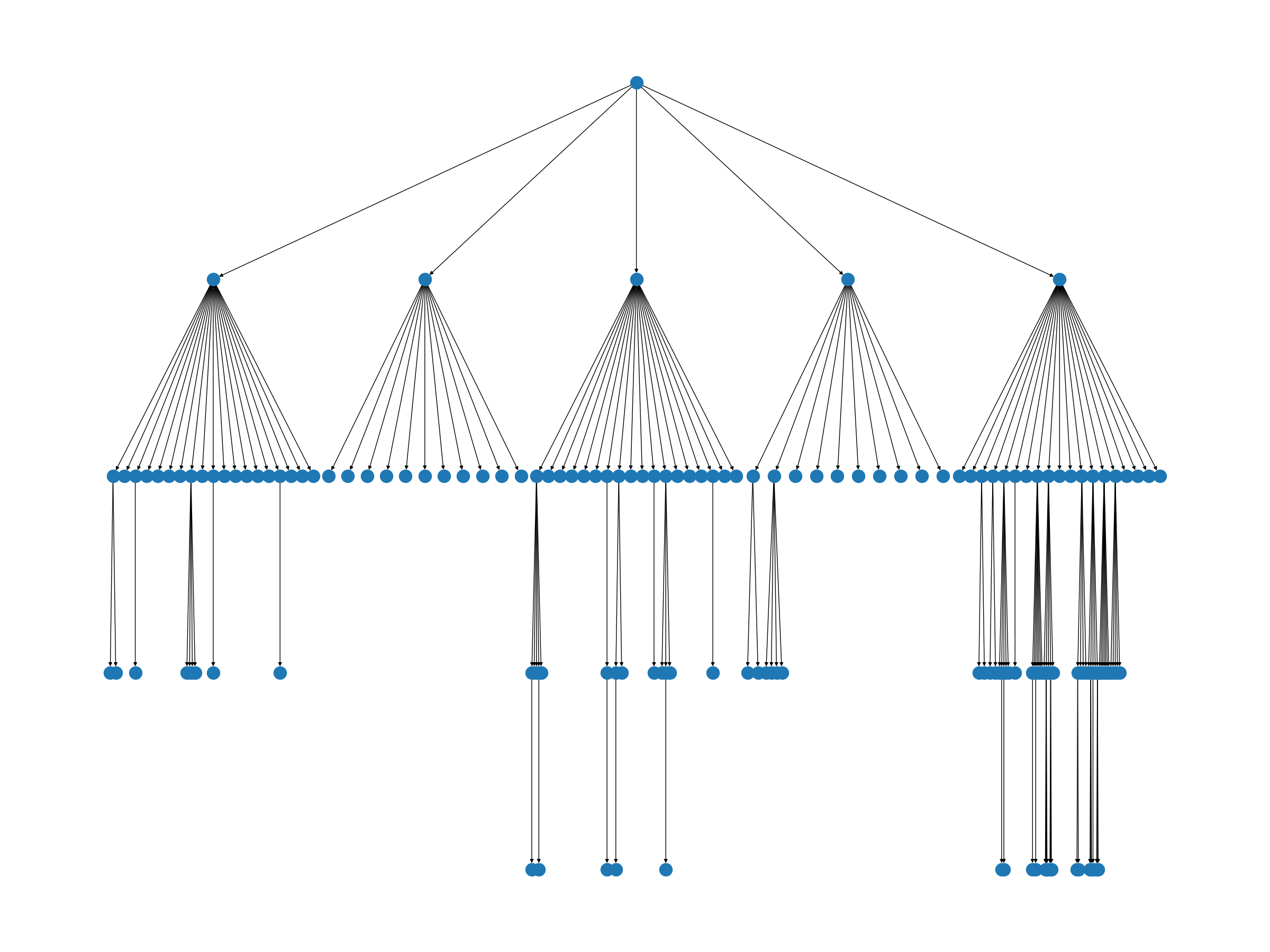}
         \caption{MAVEN}
         \label{fig:maven_ontology}
     \end{subfigure}
    \begin{subfigure}[b]{\linewidth}
         \centering
         \includegraphics[width=\linewidth]{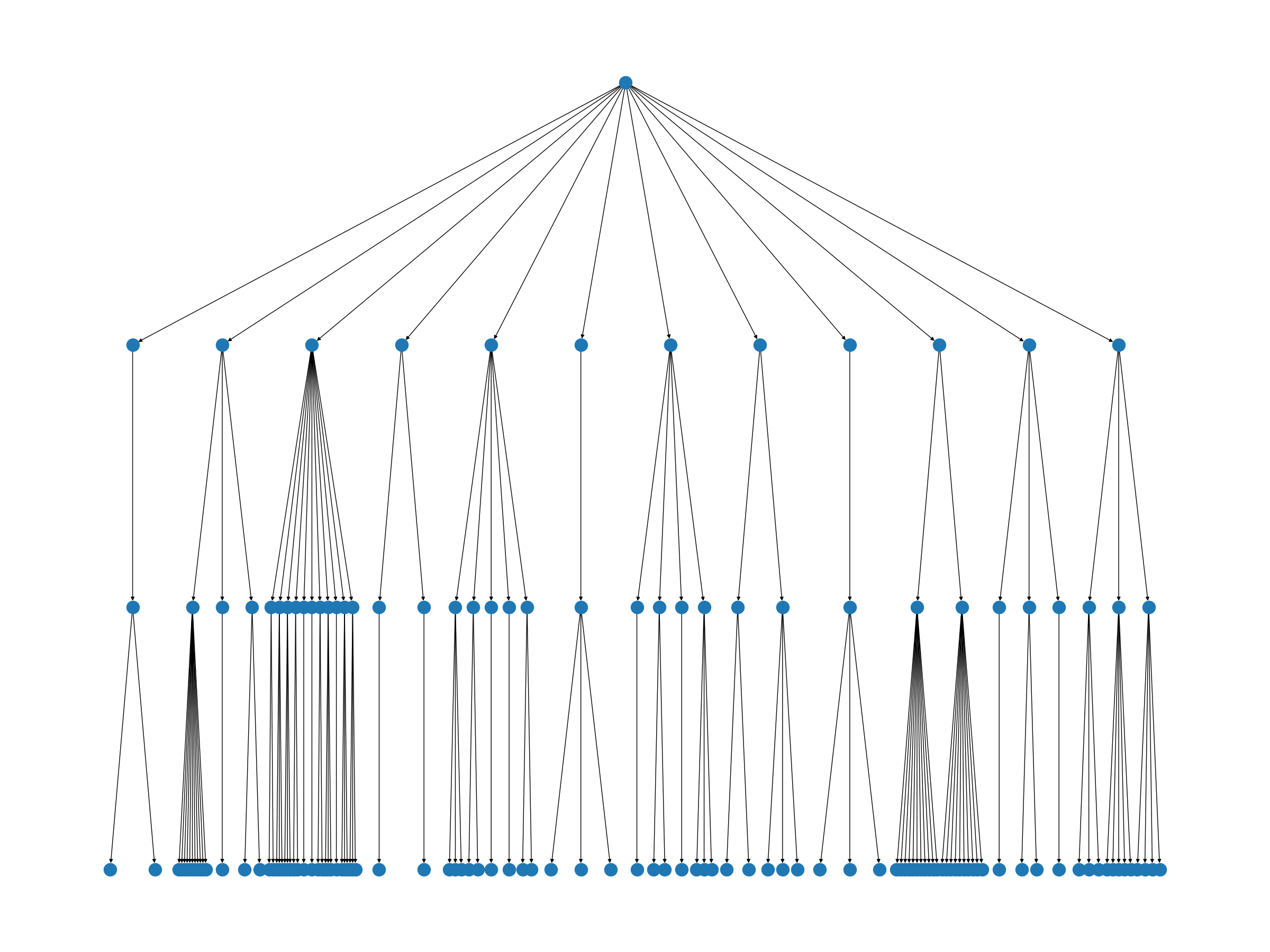}
         \caption{RAMS}
         \label{fig:rams_ontology}
     \end{subfigure}
        \caption{Event ontologies of three studied datasets.}
        \label{fig:event_ontologies}
\end{figure}

\begin{figure*}[t!]
     \centering
     \begin{subfigure}[b]{\textwidth}
         \centering
         \includegraphics[width=\textwidth]{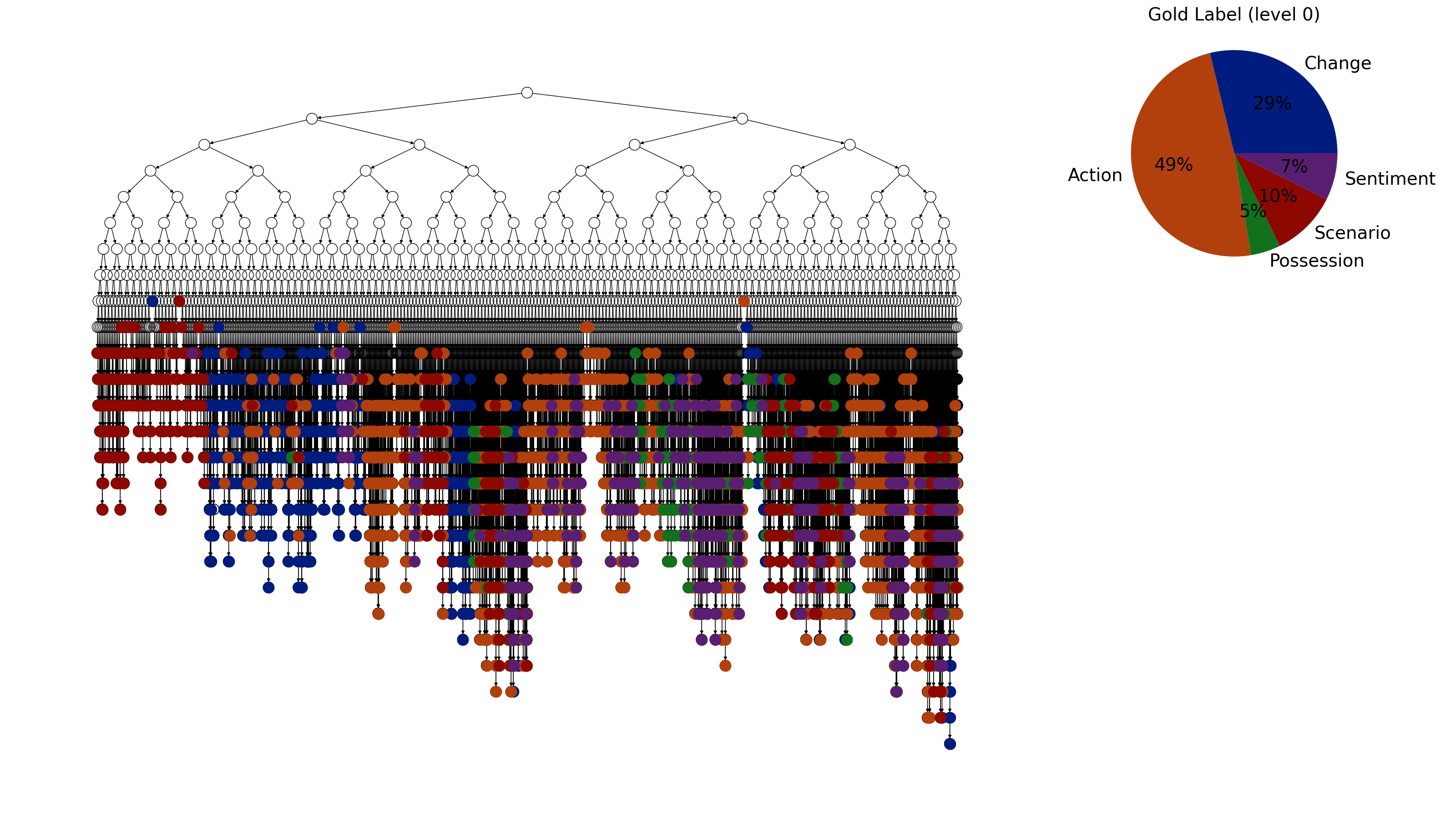}
         \caption{MAVEN}
         \label{fig:maven2005_full}
     \end{subfigure}
     \hfill
     \begin{subfigure}[b]{\textwidth}
         \centering
         \includegraphics[width=\textwidth]{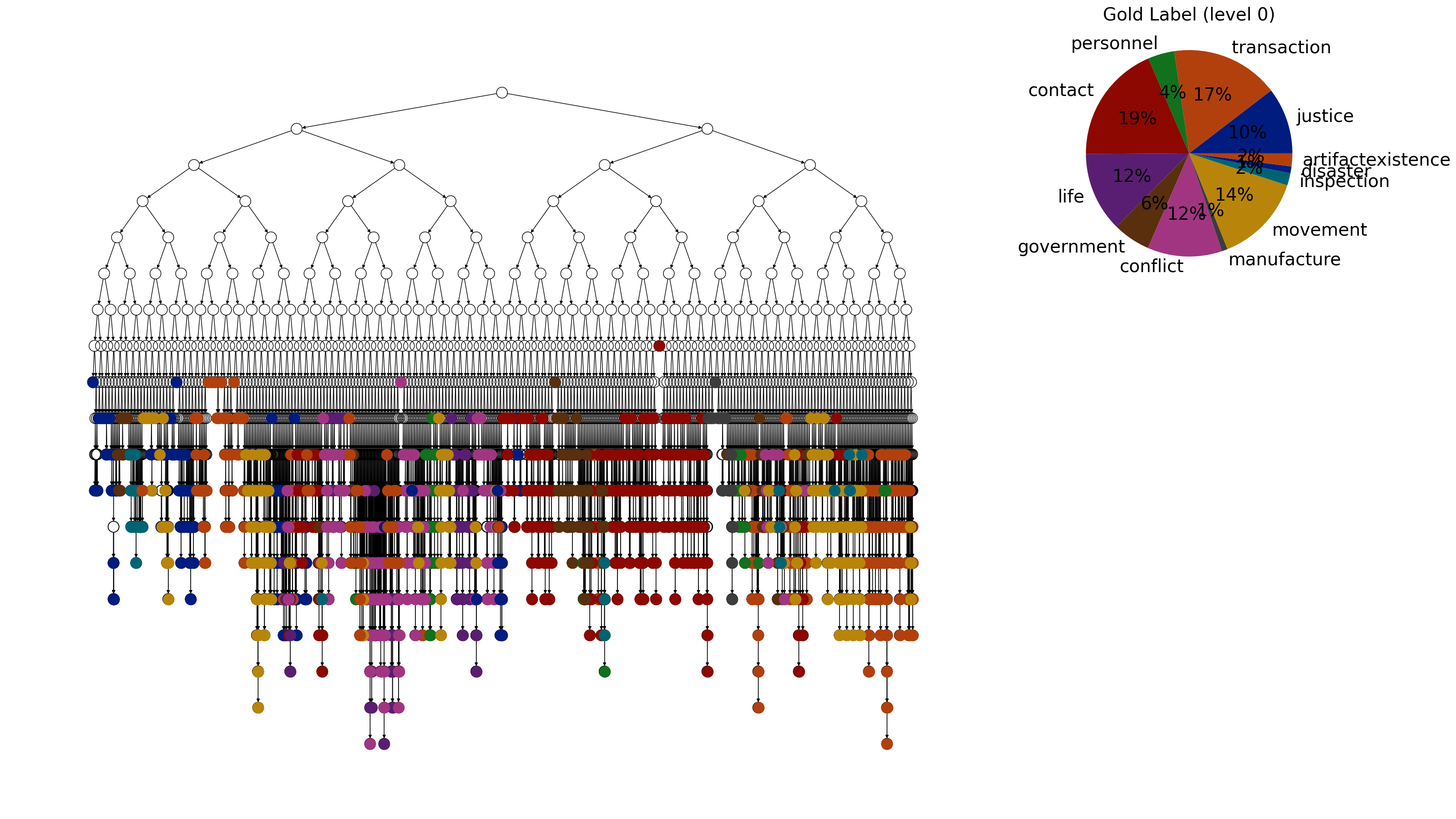}
         \caption{RAMS}
         \label{fig:rams_full}
     \end{subfigure}
        \caption{Event ontology induced by ward linkage algorithm and level-1 event type distributions on MAVEN and RAMS.}
        \label{fig:datasets_full}
\end{figure*}


\begin{table*}[t!]
\resizebox{\textwidth}{!}{%
\begin{tabular}{@{}lcccccccccc@{}}\toprule
\multirow{2}{*}{Dataset} & \multicolumn{2}{c}{spkmeans} & \multicolumn{2}{c}{kmeans} & \multicolumn{2}{c}{aggclus} & \multicolumn{2}{c}{jcsc}  & \multicolumn{2}{c}{EtypeClus} \\\cmidrule(lr){2-11}
                         & EtypeClus   & CEO            & EtypeClus      & CEO       & EtypeClus  & CEO            & EtypeClus & CEO           & EtypeClus   & CEO             \\\midrule\midrule
\multicolumn{11}{c}{BCubed\_f1}                                                                                                                                                \\
ACE2005                  & .378        & .500           & .398           & .536      & .351       & .527           & .533      & \textbf{.576} & .510        & .388            \\
MAVEN                    & .241        & .390           & .226           & .370      & .162       & \textbf{.421}  & .358      & .366          & .295        & \textbf{.395}   \\
RAMS                     & .310        & \textbf{.371}  & .302           & .359      & .306       & .380           & .380      & \textbf{.385} & .351        & .364            \\\midrule\midrule
\multicolumn{11}{c}{NMI}                                                                                                                                                       \\
ACE2005                  & .524        & .629           & .537           & .631      & .481       & .628           & .626      & \textbf{.651} & .609        & .437            \\
MAVEN                    & .522        & .676           & .503           & .663      & .428       & \textbf{.695}  & .636      & .626          & .567        & .688            \\
RAMS                     & .665        & .701           & .662           & .688      & .663       & \textbf{.706}  & .697      & .685          & .702        & .697    \\\bottomrule       
\end{tabular}%
}
\caption{Flat clustering performance of different algorithms given events represented by EtypeClus and our \textbackslash{}CEO. Higher scores indicate better clustering performance for both metrics.}
\label{tab:emb_res_others}
\end{table*}


\begin{table*}[t!]
\begin{tabular}{@{}lllllllll@{}}
\toprule
Dataset    & Method     & P@1  & P@5  & P@10  & R@1  & R@5  & R@10 & AUC  \\\midrule
NYT        & KCE~\cite{liu-etal-2018-automatic}        & .618 & .523 & 0.444 & .116 & .395 & .580 & .803 \\
           & CEE-IEA~\cite{jindal-etal-2020-killed}    & .654 & .542 & .449  & .131 & .420 & .596 & -    \\
           & \CEO & \textbf{.741} & \textbf{.604} & \textbf{.488}  & \textbf{.173} & \textbf{.493} & \textbf{.662} & \textbf{.874} \\\midrule
DailyMail  & \CEO & .438 & .309 & .316  & .169 & .491 & .639 & .753 \\\midrule
Multi-News & Longformer & .512 & .365 & .267  & .169 & .475 & .626 & .769\\\bottomrule
\end{tabular}
\caption{Salient Event Detection Performance on the test set of three datasets. The proposed \CEO fine-tunes the Longformer model to process long documents for contextualized embedding learning. It outperforms baselines with the performance reported in their papers: KCE is a kernel-based approach to learning from different statistical features, while CEE-IEA leverages token-level embeddings of all constituents from the document encoded using BERT.}
\label{tab:summary_salience_performance}
\end{table*}

\begin{figure*}[t!]
\centering
\includegraphics[width=\linewidth]{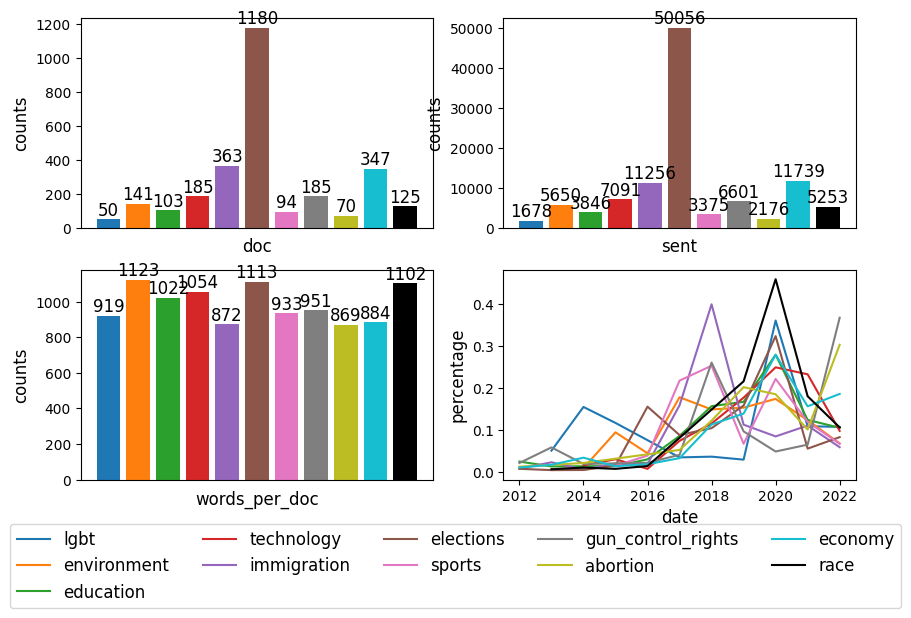}
\caption{Data statistics of the collected articles concerning 11 topics from Allsides. We record the number of documents, sentences, words per document, and distribution of released dates.}
\label{fig:allsides_statistics}
\end{figure*}

\begin{table*}[t!]
\resizebox{\textwidth}{!}{%
\begin{tabular}{@{}ll@{}}
\toprule
Topic &
  Event Instances \& Generated Names \\ \midrule
\parbox[t]{2mm}{\multirow{8}{*}{\rotatebox[origin=c]{90}{\textbf{Economy}}}}
&
  \emph{S9}: Across the nation, protesters are taking to the streets and business owners are \colorbox{yellow}{\emph{filing}} lawsuits objecting to the shutdown rules. \\
 &
  \emph{GPT-J-6B}: \textcolor{snsred}{pay:create:cause:spend:give:claim:seek} \\\cmidrule(l){2-2}
 &
  \begin{tabular}[c]{@{}l@{}}\emph{S10}: A lockdown targeted to protecting the highest-risk group, people 65 and over, instead of confining all age groups would slash deaths by\\ half but at only half the economic cost of a total \colorbox{yellow}{\emph{shutdown}}...\end{tabular} \\
 &
  \emph{GPT-J-6B}: \textcolor{snsred}{pay:create:cause:l:shut:prevent} \\\cmidrule(l){2-2}
 &
  \emph{S11}: A sharp \colorbox{yellow}{\emph{devaluation}} of the ruble would mean a drop in the standard of living for the average Russian, economists and analysts said. \\
 &
  \emph{GPT-J-6B}: \textcolor{snsred}{pay:create:cause:trade} \\\cmidrule(l){2-2}
 &
  \begin{tabular}[c]{@{}l@{}}\emph{S12}: But the NBER has other criteria that can constitute a recession, which is particularly applicable to the COVID-19 \colorbox{yellow}{\emph{crisis}} given the speed \\of the economic downturn.\end{tabular} \\
 &
  \emph{GPT-J-6B}: \textcolor{snsred}{pay:create:cause:recession:cat:crisis} \\\midrule
\parbox[t]{2mm}{\multirow{8}{*}{\rotatebox[origin=c]{90}{\textbf{Education}}}}
&
  \begin{tabular}[c]{@{}l@{}}\emph{S13}: On July 28, the American Federation of Teachers, the second-largest education \colorbox{yellow}{\emph{union}}, threatened "safety strikes" if reopening plans aren't\\ entirely to its liking.\end{tabular} \\
 &
  \emph{GPT-J-6B}: \textcolor{snsred}{pay:education:teach:organ:organization} \\\cmidrule(l){2-2}
 &
  \emph{S14}: ...Obama said during an online commencement address to \colorbox{yellow}{\emph{graduates}} of historically black colleges and universities (HBCUs) on Saturday. \\
 &
  \emph{GPT-J-6B}: \textcolor{snsred}{pay:education:get} \\\cmidrule(l){2-2}
 &
  \begin{tabular}[c]{@{}l@{}}\emph{S15}: ...a conspiracy theory pushed by the president that accuses Obama of attempting to frame Trump for colluding with Russia to win the 2016\\ \colorbox{yellow}{\emph{election}}.\end{tabular} \\
 &
  \emph{GPT-J-6B}: \textcolor{snsred}{pay:education:cause:app:vote:election} \\\cmidrule(l){2-2}
 &
  \emph{S16}: Yet ... six of them carry the \colorbox{yellow}{\emph{support}} of more than 50 percent of committed liberals ... \\
 &
  \emph{GPT-J-6B}: \textcolor{snsred}{pay:education:cause:enjoy:support} \\ \midrule 
\parbox[t]{2mm}{\multirow{8}{*}{\rotatebox[origin=c]{90}{\textbf{Environment}}}}
&
  \emph{S17}: Satellite data published by the National Institute for Space research (Inpe) shows an increase of 85\% this year in \colorbox{yellow}{\emph{fires}} across Brazil... \\
 &
  \emph{GPT-J-6B}: \textcolor{snsred}{be:cause:burn} \\\cmidrule(l){2-2}
 &
  \begin{tabular}[c]{@{}l@{}}\emph{S18}: Indeed, when the scientists drew up their first \colorbox{yellow}{\emph{report}}, in 1990, the diplomats tried so hard to water down their conclusions that the  whole \\enterprise nearly collapsed.\end{tabular} \\
 &
  \emph{GPT-J-6B}: \textcolor{snsred}{be:cause:report:find:release} \\\cmidrule(l){2-2}
 &
  \begin{tabular}[c]{@{}l@{}}\emph{S19}: It is likely going to make the world sicker, hungrier, poorer, gloomier and way more dangerous in the next 18 years with an "unavoidable" \\\colorbox{yellow}{\emph{increase}} in risks...\end{tabular} \\
 &
  \emph{GPT-J-6B}: \textcolor{snsred}{be:cause:make:change:reduce:growth:increase} \\\cmidrule(l){2-2}
 &
  \begin{tabular}[c]{@{}l@{}}\emph{S20}: Supporters of Mr. Obama's \colorbox{yellow}{\emph{plan}}, including some Democratic-led states and environmental groups, argue it will create thousands of clean\\-energy jobs and help...\end{tabular} \\
 &
  \emph{GPT-J-6B}: \textcolor{snsred}{be:cause:policy:plan} \\\midrule
\parbox[t]{2mm}{\multirow{8}{*}{\rotatebox[origin=c]{90}{\textbf{Gun Control Rights}}}}
&
  \begin{tabular}[c]{@{}l@{}}\emph{S21}: LaPierre told Friday's audience "every NRA member is in mourning" because of the Uvalde \colorbox{yellow}{\emph{shooting}}, which he said was the work of a \\"criminal monster."\end{tabular} \\
 &
  \emph{GPT-J-6B}: \textcolor{snsred}{kill:shoot} \\\cmidrule(l){2-2}
 &
  \emph{S22}: ...Houston and the gun \colorbox{yellow}{\emph{safety}} group Moms Demand Action, held protests outside the convention center Friday. \\
 &
  \emph{GPT-J-6B}: \textcolor{snsred}{kill:control:make:cause:safety} \\\cmidrule(l){2-2}
 &
  \begin{tabular}[c]{@{}l@{}}\emph{S23}: Mr. Biden also \colorbox{yellow}{\emph{urged}} lawmakers to expand background checks for gun purchases, change liability laws to allow gun manufacturers to be \\sued for shootings...\end{tabular} \\
 &
  \emph{GPT-J-6B}: \textcolor{snsred}{kill:control:make:cause:protest:spend:motion:closing:request} \\\cmidrule(l){2-2}
 &
  \begin{tabular}[c]{@{}l@{}}\emph{S24}: It would raise the federal age of purchasing a rifle from 18 to 21; \colorbox{yellow}{\emph{restrict}} ammunition magazine capacity, though existing magazines are \\"grandfathered" in...\end{tabular} \\
 &
  \emph{GPT-J-6B}: kill:control:make:ban:restrict \\\midrule
\parbox[t]{2mm}{\multirow{8}{*}{\rotatebox[origin=c]{90}{\textbf{Immigration}}}}
&
  \emph{S25}: There were \colorbox{yellow}{\emph{immigrants}} from El Salvador, China, Honduras and countries in between. \\
 &
  \emph{GPT-J-6B}: \textcolor{snsred}{cause:imigration} \\\cmidrule(l){2-2}
 &
  \begin{tabular}[c]{@{}l@{}}\emph{S26}: ...She spoke the same night President Trump in a message on Twitter said that Immigration and Customs Enforcement next week would\\ begin \colorbox{yellow}{\emph{deporting}} "millions"  of immigrants who are living in the U.S. illegally.\end{tabular} \\
 &
  \emph{GPT-J-6B}: \textcolor{snsred}{cause:immigration:death:travel:seek:arrest:hold:removal} \\\cmidrule(l){2-2}
 &
  \begin{tabular}[c]{@{}l@{}}\emph{S27}: Democrats are likely to face questions about whether they agree with Ocasio-Cortez's comments about concentration camps and the\\ Trump administration's  \colorbox{yellow}{\emph{detention}} centers as they return to Washington this week.\end{tabular} \\
 &
  \emph{GPT-J-6B}: \textcolor{snsred}{cause:immigration:death:travel:seek:arrest:hold} \\\cmidrule(l){2-2}
 &
  \begin{tabular}[c]{@{}l@{}}\emph{S28}: ... progressives and Democratic congressional leaders have been pressuring Biden to \colorbox{yellow}{\emph{end}} the use of the policy that turns back families\\ and single adults at the border.\end{tabular} \\
 &
  \emph{GPT-J-6B}: \textcolor{snsred}{cause:closing:end:process} \\\bottomrule
  \end{tabular}%
}
\caption{Identified events and generated type names for instances sampled from 5 topics of Allsides.}
\label{tab:allsides_more_examples_5}
\end{table*}

\begin{table*}[t!]
\resizebox{\textwidth}{!}{%
\begin{tabular}{@{}ll@{}}
\toprule
Topic &
  Event Instances \& Generated Names \\ \midrule
\parbox[t]{2mm}{\multirow{8}{*}{\rotatebox[origin=c]{90}{\textbf{Elections}}}}
&
  \emph{S29}: That's consonant with broad \colorbox{yellow}{\emph{support}} for police generally. \\
 &
  \emph{GPT-J-6B}: \textcolor{snsred}{election:debate:cause:support} \\\cmidrule(l){2-2}
 &
  \emph{S30}: A number of prominent figures have explicitly \colorbox{yellow}{\emph{called}} for defunding or abolition of police. \\
 &
  \emph{GPT-J-6B}: \textcolor{snsred}{election:win:be:think:make:call} \\\cmidrule(l){2-2}
 &
  \begin{tabular}[c]{@{}l@{}}\emph{S31}: A majority of members of the City Council of Minneapolis... \colorbox{yellow}{\emph{announced}} over the weekend their plans to "begin the process of \\ ending the Minneapolis Police Department."\end{tabular} \\
 &
  \emph{GPT-J-6B}: \textcolor{snsred}{election:debate:cause:support:end:announce:campaign} \\\cmidrule(l){2-2}
 &
  \begin{tabular}[c]{@{}l@{}}\emph{S32}: ...Democratic presidential candidate Joe Biden said Monday he \colorbox{yellow}{\emph{opposes}} "defunding the police," declining to embrace a rallying cry \\that has gained support...\end{tabular} \\
 &
  \emph{GPT-J-6B}: \textcolor{snsred}{election:debate:cause:support:attack:contest:opposition} \\\midrule
\parbox[t]{2mm}{\multirow{8}{*}{\rotatebox[origin=c]{90}{\textbf{Race}}}} &
  \emph{S33}: In San Francisco, the mob \colorbox{yellow}{\emph{demolished}} statues of Ulysses S. Grant, Junipero Serra, and Francis Scott Key. \\
 &
  \emph{GPT-J-6B}: \textcolor{snsred}{kill:cause:protest:crit:ban:celebr:end:destruction} \\\cmidrule(l){2-2}
 &
  \emph{S34}: Last week a mob in downtown Washington, D.C. decided to \colorbox{yellow}{\emph{tear}} down a statue of a man called Albert Pike. \\
 &
  \emph{GPT-J-6B}: \textcolor{snsred}{kill:be:cause:removal:destruction:t} \\\cmidrule(l){2-2}
 &
  \emph{S35}: This is a serious and highly organized political \colorbox{yellow}{\emph{movement}}. \\
 &
  \emph{GPT-J-6B}: \textcolor{snsred}{kill:be:cause:give:host:protest} \\\cmidrule(l){2-2}
 &
  \begin{tabular}[c]{@{}l@{}}\emph{S36}: \colorbox{yellow}{\emph{Reforms}} have also been proposed under "8 Can't Wait," an initiative released in the wake of the protests by Campaign Zero, a group \\advocating police reform.\end{tabular} \\
 &
  \emph{GPT-J-6B}: \textcolor{snsred}{kill:cause:death:process:reform} \\\midrule
\parbox[t]{2mm}{\multirow{8}{*}{\rotatebox[origin=c]{90}{\textbf{Sports}}}}
&
  \begin{tabular}[c]{@{}l@{}}\emph{S37}: The United States \colorbox{yellow}{\emph{beat}} the Netherlands in the 2019 Women's World Cup on Sunday 2-0, following a month-long tournament that\\ attracted more attention to the sport...\end{tabular} \\
 &
  \emph{GPT-J-6B}: \textcolor{snsred}{protest:be:watch:give:win} \\\cmidrule(l){2-2}
 &
  \begin{tabular}[c]{@{}l@{}}\emph{S38}: After other hits including "Earned It" and "Save Your Tears,"The Weeknd concluded the 13-minute \colorbox{yellow}{\emph{show}} with his smash single \\ "Blinding Lights," a song that references...\end{tabular} \\
 &
  \emph{GPT-J-6B}: \textcolor{snsred}{protest:advertising:cause:give:meet:view:coverage:performance} \\\cmidrule(l){2-2}
 &
  \begin{tabular}[c]{@{}l@{}}\emph{S39}: But this year, many advertising insiders \colorbox{yellow}{\emph{expect}} the Super Bowl spots to steer clear of the \#MeToo movement opposing the sexual\\ harassment and abuse of women...\end{tabular} \\
 &
  \emph{GPT-J-6B}: \textcolor{snsred}{protest:be:watch:give:agreement:predict} \\\cmidrule(l){2-2}
 &
  \emph{S40}: ...city councils, governors and state legislatures all too often respond by \colorbox{yellow}{\emph{offering}} lucrative "inducement payments." \\
 &
  \emph{GPT-J-6B}: \textcolor{snsred}{protest:be:watch:give} \\\midrule
\parbox[t]{2mm}{\multirow{8}{*}{\rotatebox[origin=c]{90}{\textbf{Technology}}}}&
  \emph{S41}: Moreno accused Assange of behaving badly at the embassy, interfering with building security and attempting to \colorbox{yellow}{\emph{access}} security files. \\
 &
  \emph{GPT-J-6B}: \textcolor{snsred}{cause:communication:service:access} \\\cmidrule(l){2-2}
 &
  \begin{tabular}[c]{@{}l@{}}\emph{S42}: "When users \colorbox{yellow}{\emph{violate}} these policies repeatedly, like our policies against hate speech and harassment or our terms prohibiting \\ circumvention of our enforcement measures...\end{tabular} \\
 &
  \emph{GPT-J-6B}: \textcolor{snsred}{cause:ban:repe:cance:break:removal} \\\cmidrule(l){2-2}
 &
  \emph{S43}: The InfoWars broadcaster's past tweets will, however, \colorbox{yellow}{\emph{remain}} viewable to others while his account is locked in a "read-only" mode. \\
 &
  \emph{GPT-J-6B}: \textcolor{snsred}{cause:control:keep:be:hold} \\\cmidrule(l){2-2}
 &
  \begin{tabular}[c]{@{}l@{}}\emph{S44}: Mr Jones subsequently \colorbox{yellow}{\emph{posted}} a video in which he discusses the move to a separate @Infowars feed - with about 431,000 followers \\ - which he described as being a "sub-account".\end{tabular} \\
 &
  \emph{GPT-J-6B}: \textcolor{snsred}{cause:publish:question:post}\\\bottomrule
\end{tabular}%
}
\caption{Identified events and generated type names for instances sampled from 4 topics of Allsides.}
\label{tab:allsides_more_examples_4}
\end{table*}

\end{document}